\documentclass[10pt,twocolumn,letterpaper]{article}

 \usepackage[pagenumbers]{cvpr} 

\usepackage{graphicx}
\usepackage{amsmath}
\usepackage{amssymb}
\usepackage{booktabs}

\usepackage{bm}
\usepackage{nicefrac}

\let\originalleft\left
\let\originalright\right
\def\left#1{\mathopen{}\originalleft#1}
\def\right#1{\originalright#1\mathclose{}}

\newcommand{\smallsim}{\smallsym{\mathrel}{\sim}}

\makeatletter
\newcommand{\smallsym}[2]{#1{\mathpalette\make@small@sym{#2}}}
\newcommand{\make@small@sym}[2]{%
  \vcenter{\hbox{$\m@th\downgrade@style#1#2$}}%
}
\newcommand{\downgrade@style}[1]{%
  \ifx#1\displaystyle\scriptstyle\else
    \ifx#1\textstyle\scriptstyle\else
      \scriptscriptstyle
  \fi\fi
}
\makeatother

\usepackage[pagebackref,breaklinks,colorlinks]{hyperref}

\usepackage[capitalize]{cleveref}
\crefname{section}{Sec.}{Secs.}
\Crefname{section}{Section}{Sections}
\Crefname{table}{Table}{Tables}
\crefname{table}{Tab.}{Tabs.}

\def\cvprPaperID{2085} 
\def\confName{CVPR}
\def\confYear{2023}

\begin{document}

\title{Shakes on a Plane: Unsupervised Depth Estimation\\ from Unstabilized Photography}


\author{Ilya Chugunov\quad Yuxuan Zhang\quad Felix Heide \vspace{0.8em}\\
Princeton University}

\maketitle

\begin{abstract}
Modern mobile burst photography pipelines capture and merge a short sequence of frames to recover an enhanced image, but often disregard the 3D nature of the scene they capture, treating pixel motion between images as a 2D aggregation problem. We show that in a ``long-burst'', forty-two 12-megapixel RAW frames captured in a two-second sequence, there is enough parallax information from natural hand tremor alone to recover high-quality scene depth. To this end, we devise a test-time optimization approach that fits a neural RGB-D representation to long-burst data and simultaneously estimates scene depth and camera motion. Our plane plus depth model is trained end-to-end, and performs coarse-to-fine refinement by controlling which multi-resolution volume features the network has access to at what time during training. We validate the method experimentally, and demonstrate geometrically accurate depth reconstructions with no additional hardware or separate data pre-processing and pose-estimation steps.
\end{abstract}

 \vspace{-1em}
\section{Introduction}
Over the last century we saw not only the rise and fall in popularity of film and DSLR photography, but of standalone cameras themselves. We've moved into an era of ubiquitous multi-sensor, multi-core, multi-use, mobile-imaging platforms~\cite{delbracio2021mobile}. Modern cellphones offer double-digit megapixel image streams at high framerates; optical image stabilization; on-board motion measurement devices such as accelerometers, gyroscopes, and magnetometers; and, most recently, integrated active depth sensors~\cite{luetzenburg2021evaluation}. This latest addition speaks to a parallel boom in the field of depth imaging and 3D reconstruction~\cite{han2019image,zollhofer2018state}. As users often photograph people, plants, food items, and other complex 3D shapes, depth can play a key role in object understanding tasks such as detection, segmentation, and tracking~\cite{ji2021calibrated,schwarz2018rgb,yang2022RGBD}. 3D information can also help compensate for non-ideal camera hardware and imaging settings through scene relighting~\cite{pandey2021total,guo2019relightables,yang2021s3net}, simulated depth-of-field effects~\cite{wadhwa2018synthetic,abuolaim2020defocus, wang2018deeplens}, and frame interpolation~\cite{bao2019depth}. Beyond helping improve or understand RGB content, depth itself is a valuable output for simulating objects in augmented reality~\cite{serrano2019motion,bertel2020omniphotos,luo2020consistent,du2020depthlab} and interactive experiences~\cite{hedman2018instant,kopf2020one}.

\begin{figure}[t]
    \centering
    \includegraphics[width=\linewidth]{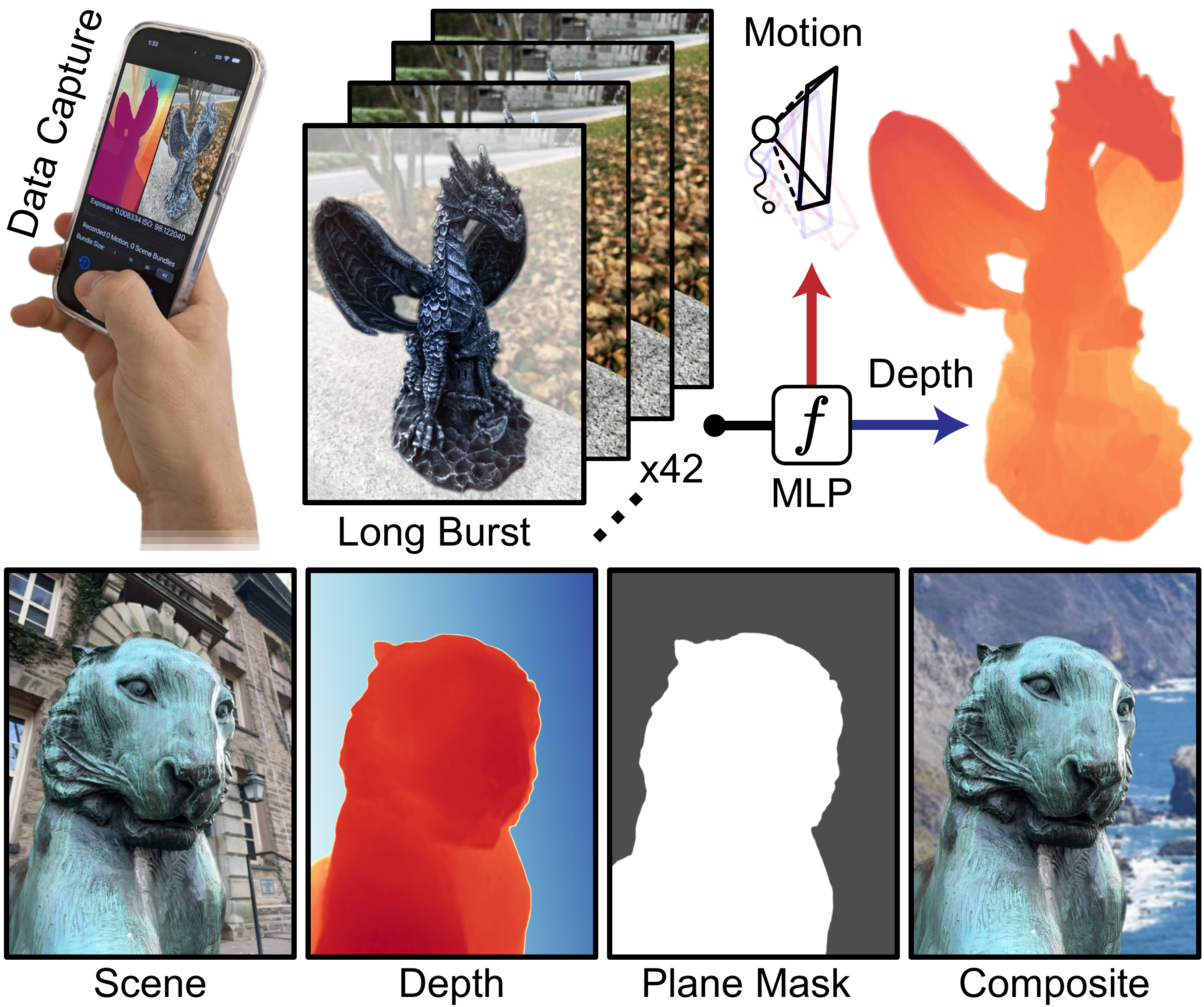}
    \caption{Our neural RGB-D model fits to a single \emph{long-burst} image stack to distill high quality depth and camera motion. The model's \emph{depth-on-a-plane} decomposition can facilitate easy background masking, segmentation, and image compositing.}
    \label{fig:my_label}
    \vspace{-1.5em}
\end{figure}

Depth reconstruction can be broadly divided into \emph{passive} and \emph{active} approaches. \emph{Passive} monocular depth estimation methods leverage training data to learn shape priors~\cite{ranftl2021vision, hu2019revisiting, bhat2021adabins} -- e.g., what image features imply curved versus flat objects or occluding versus occluded structures -- but have a hard time generalizing to out-of-distribution scenes~\cite{ranftl2020towards,miangoleh2021boosting}. Multi-view depth estimation methods lower this dependence on learned priors by leveraging parallax information from camera motion~\cite{fonder2021m4depth, ummenhofer2017demon} or multiple cameras~\cite{tankovich2021hitnet, marr1976cooperative} to recover geometrically-guided depth. The recent explosion in neural radiance field approaches~\cite{mildenhall2020nerf, yu2021pixelnerf, tancik2022block, mildenhall2022nerf} can be seen a branch of multi-view stereo where a system of explicit geometric constraints is swapped for a more general learned scene model. Rather than classic feature extraction and matching, these models are fit directly to image data to distill dense \textit{implicit} 3D information.

 \emph{Active} depth methods such as pulsed time-of-flight~\cite{mccarthy2009long} (e.g., LiDAR), correlation time-of-flight~\cite{lange20003d}, and structured light~\cite{zhang2018high, scharstein2003high} use \emph{controlled illumination} to help with depth reconstruction. While these methods are less reliant on image content than \emph{passive} ones, they also come with complex circuitry and increased power demands~\cite{horaud2016overview}. Thus, miniaturization for mobile applications results in very low-resolution \emph{sub-kilopixel sensors}~\cite{hegblom2022column, warren2018low,callenberg2021low}. The Apple iPhone 12-14 Pro devices, which feature one of these miniaturized sensors, use depth derived from RGB, available at 12 \emph{mega-pixel} resolution, to recover scene details lost in the sparse LiDAR measurements. While how exactly they use the RGB stream is unknown, occluding camera sensors reveals that the estimated geometry is the result of \emph{monocular} RGB-guided depth reconstruction.

Returning to the context of mobile imaging, even several seconds of continuous mode photography, which we refer to as a ``long-burst", contain only millimeter-scale view variation from natural hand tremor~\cite{chugunov2022implicit}. While these \emph{micro-baseline}~\cite{joshi2014micro} shifts are effectively used in burst superresolution and denoising methods~\cite{wronski2019handheld, qian2019rethinking} as indirect observations of content between sensor pixels, 3D parallax effects on pixel motion are commonly ignored in these models as the depth recovered from this data is too coarse for sub-pixel refinement~\cite{yu20143d, joshi2014micro,im2015high}. A recent work~\cite{chugunov2022implicit} demonstrates high-quality object reconstructions refined with long-burst RGB data, but relies on the iPhone 12 Pro LiDAR sensor for initial depth estimates and device poses, not available on many other cellphones. They treat these poses as ground truth and explicitly solve for depth through minimization of photometric reprojection loss.

In this work, we devise an unsupervised end-to-end approach to jointly estimate high-quality object depth and camera motion from more easily attainable unstabilized two-second captures of 12-megapixel RAW frames and gyroscope data. Our method requires no depth initialization or pose inputs, only a long-burst. We formulate the problem as an image synthesis task, similar to neural radiance methods~\cite{mildenhall2020nerf}, decomposed into explicit geometric projection through continuous depth and pose models. In contrast to recent neural radiance methods, which typically estimate poses in a pre-processing step, we jointly distill relative depth and pose estimates as a product of simply fitting our model to long-burst data and minimizing photometric loss. In summary, we make the following contributions:
\begin{itemize}
    \item An end-to-end neural RGB-D scene fitting approach that distills high-fidelity affine depth and camera pose estimates from unstabilized long-burst photography.\vspace{-0.25em}
    \item A smartphone data collection application to capture RAW images, camera intrinsics, and gyroscope data for our method, as well as processed RGB frames, low-resolution depth maps, and other camera metadata.    
    \item Evaluations which demonstrate that our approach outperforms existing single and multi-frame image-only depth estimation approaches, with comparisons to high-precision structured light scans to validate the accuracy of our reconstructed object geometries.
\end{itemize}
Code, data, videos, and additional materials are available on our project website: \href{https://light.princeton.edu/soap/}{https://light.princeton.edu/soap}

\section{Related Work} 
There exist a wide array of both \emph{active} and \emph{passive} depth estimation methods: ones that recover depth with the help of a controlled illumination source, and ones that use only naturally collected light. We review related work in both categories before discussing neural scene representations.

\vspace{1em}
\noindent\textbf{Active Depth Reconstruction.}\hspace{0.1em} Structured light and active stereo methods rely on patterned illumination to directly infer object shape~\cite{zhang2018high, fanello2016hyperdepth} and/or improve stereo feature matching~\cite{scharstein2003high}. In contrast, time-of-flight (ToF) depth sensors use the round trip time of photons themselves -- how long it takes light to reach and return from an object -- to infer depth. \emph{Indirect} ToF does this by calculating phase changes in continuously modulated light~\cite{lange20003d,hansard2012time,kolb2010time}, while \emph{direct} ToF times how long a pulse of light is in flight to estimate depth~\cite{morimoto2020megapixel,mccarthy2009long}. The LiDAR system found in the iPhone 12-14 Pro devices is a type of direct ToF sensor built on low-cost single-photon detectors~\cite{callenberg2021low} and solid-state vertical-cavity surface-emitting laser technology~\cite{warren2018low}. While active LiDAR depth measurements can help produce \emph{metric} depth estimates -- without scale ambiguity -- existing mobile depth sensors have very limited sub-kilopixel spatial resolution, are sensitive to surface reflectance, and are not commonly found on other mobile devices.

\vspace{1em}
\noindent\textbf{Passive Depth Reconstruction.}\hspace{0.1em} Single-image passive methods leverage the correlation between visual and geometric features to estimate 3D structure. Examples include depth from shading~\cite{barron2014shape, xiong2014shading}, focus cues~\cite{xiong1993depth}, or generic learned priors~\cite{ranftl2021vision, hu2019revisiting, bhat2021adabins}. Learned methods have shown great success in producing visually coherent results, but rely heavily on labeled training data and produce unpredictable outputs for out-of-distribution samples. Multi-view and structure from motion works leverage epipolar geometry~\cite{hartley2003multiple}, the relationship between camera and image motion, to extract 3D information from multiple images. Methods typically either directly match RGB features~\cite{sinha2007multi,galliani2015massively}, or higher-level learned features~\cite{tankovich2021hitnet,lipson2021raft}, in search of depth and/or camera parameters which maximize \emph{photometric consistency} between frames. COLMAP~\cite{schonberger2016structure} is a widely adopted multi-view method which many neural radiance works~\cite{mildenhall2020nerf, mildenhall2022nerf} rely on for camera pose estimates. In the case of long-burst photography, this problem becomes significantly more challenging as many different depth solutions produce identical images under small view variations. Work in this space either relies on interpolation between sparse feature matches~\cite{yu20143d, joshi2014micro,im2015high} or additional hardware~\cite{chugunov2022implicit}  to produce complete depth estimates. Our work builds on these methods to produce both dense depth and high-accuracy camera motion estimates from long-burst image data alone, with a single model trained end-to-end rather than a sequence of disjoint data processing steps.

\vspace{1em}
\noindent\textbf{Neural Scene Representations.}\hspace{0.1em}
Recent work in the area of novel view synthesis has demonstrated that explicit models -- e.g. voxel grids, point clouds, or depth maps -- are not a necessary backbone to generate high-fidelity representations of 3D space. Rather, the neural radiance family of works, including NeRF~\cite{mildenhall2020nerf} and its extensions \cite{chen2021mvsnerf, barron2021mip, muller2022instant}, learn an implicit representation of a 3D scene by fitting a multi-layer perceptron (an MLP)~\cite{hornik1989multilayer} to a set of input images through gradient descent. Similar to multi-view stereo, these methods optimize for photometric loss, ensuring output colors match the underlying RGB data, but they typically don't produce depth maps or camera poses as outputs. On the contrary, most neural radiance methods require camera poses as inputs, obtained from COLMAP~\cite{schonberger2016structure} in a separate pre-processing step. Our setting of long-burst unstabilized photography not only lacks ground truth camera poses, but also provides very little view variation from which to estimate them. While neural scene representation works exist which learn camera poses~\cite{lin2021barf, wang2021nerf}, or operate in the burst photography setting~\cite{pearl2022nan}, to our best knowledge this is the first work to jointly do both. The most similar recent work by Chugunov et al.~\cite{chugunov2022implicit} uses poses derived from the iPhone 12 Pro ARKit library to learn an implicit representation of depth, but \emph{does not have an image generation model}, and is functionally closer to a direct multi-view stereo approach. In contrast, our work uses a neural representation of RGB as an optimization vehicle to distill high quality continuous representations of both depth and camera poses, with loss backpropogated through an explicit 3D projection model.

\section{Long-Burst Photography}
\begin{figure}[t]
    \centering
    \includegraphics[width=\linewidth]{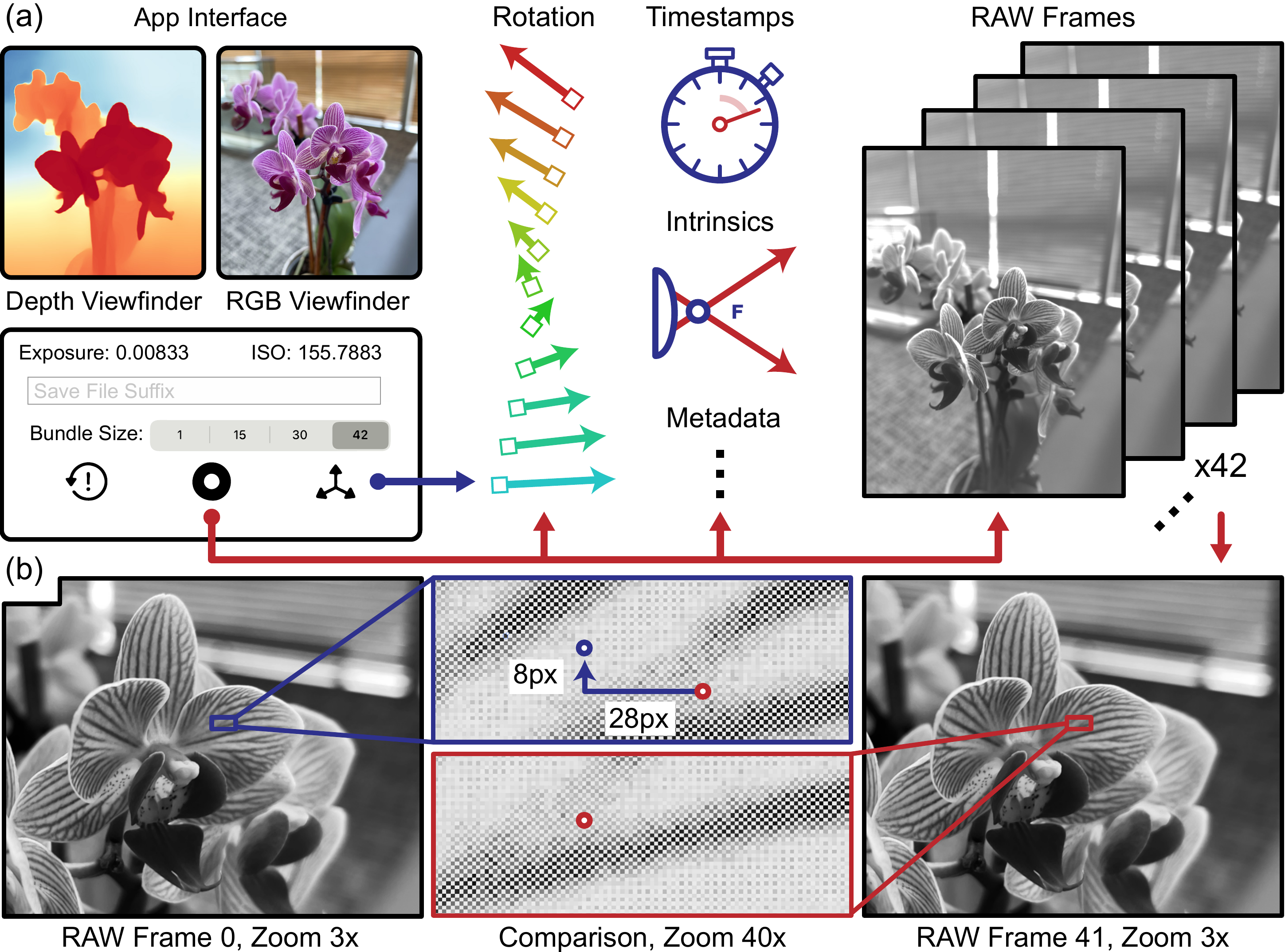}
    \caption{ (a) The interface of our app for recording long-burst data. (b) Aligned RAW frames, which illustrate the scale of parallax motion created by natural hand tremor in a two-second capture: a few dozen pixels for an object 30cm from the camera. }
    \label{fig:app}
    \vspace{-1.5em}
\end{figure}

\noindent\textbf{Problem Setting.}\hspace{0.1em} Burst photography refers to the imaging setting where for each button press from the user the camera records multiple frames in rapid succession, sometimes varying parameters such as ISO and exposure time during capture to create a \emph{bracketed sequence}~\cite{mertens2009exposure}. Burst imaging pipelines investigate how these frames can be merged back into a single higher-fidelity image~\cite{delbracio2021mobile}. These pipelines \emph{typically operate with 2-8 frame captures} and have proven key to high-quality mobile imaging in low-light~\cite{liba2019handheld, hasinoff2016burst}, high dynamic range imaging with low dynamic range sensors~\cite{hasinoff2016burst,gallo2015locally}, and image superresolution, demosaicing, and denoising~\cite{wronski2019handheld, weissman2005universal}. On the other end of the imaging spectrum we have video processing literature, which operates on sequences hundreds or thousands of frames in length~\cite{monakhova2022dancing} and/or with large camera motion~\cite{kopf2021robust}. Between these two settings we have what we refer to as ``long-burst" photography, several seconds of continuous capture with small view variation. Features built into default mobile camera applications such as Android Motion Photos and Apple Live Photos, which both record three seconds of frames around a button press, demonstrate the ubiquity of \emph{long-burst} data. They are captured spontaneously, without user interaction, during natural handheld photography. In this work we capture two-second long-bursts, which result in 42 recorded frames with an average 6mm maximum effective stereo baseline. As seen in Fig.~\ref{fig:app} (b), this produces on the order of several dozen pixels of disparity for close-range objects ($<$0.5m). For an in-depth discussion of motion from natural hand tremor we refer the reader to Chugunov et al.~\cite{chugunov2022implicit}.

\noindent\textbf{Data Collection.}\hspace{0.1em} As there were no commodity mobile applications that allowed for continuous streaming of Bayer RAW frames and metadata, we designed our own data collection tool for long-burst recording. Shown in Fig.~\ref{fig:app} (a), it features a live viewfinder with previews of RGB, device depth, and auto-adjusted ISO and exposure values. On a button press, we lock ISO, exposure, and focus, and record a two-second, 42 frame long-burst to the device. Our method uses recorded timestamps, camera intrinsics, gyroscope-driven device rotation estimates, and 12-megapixel RAW frames. However, our app also records processed RGB frames, low-resolution depth maps, and other metadata which we use for validation and visualization. 

\noindent\textbf{RAW Images.\footnote{\emph{RAW} here refers to sensor data after basic corrections such as compensating for broken and non-uniform pixels, not ``raw-raw'' data~\cite{pulli_2022}.}}\hspace{0.1em} A modern mobile image signal processing pipeline can have more than a dozen steps between light hitting the CMOS sensor and a photo appearing on screen: denoising, demosaicing, and gamma correction to name a few~\cite{delbracio2021mobile}. While these steps, when finely-tuned, can produce eye-pleasing results, they also pose a problem to down-stream computer vision tasks: they break linear noise assumptions, correlate pixel neighborhoods, and lower the overall dynamic range of the content (quantizing the 10- to 14-bit sensor measurements down to 8-bit color depth image files)~\cite{brooks2019unprocessing}.
In our work we are concerned with the tracking and reconstruction of small image features undergoing small continuous motion from natural hand tremor, and so apply minimal processing to our image data, using linear interpolation to only fill the gaps between Bayer measurements. We preserve the full 14-bit color depth, and fit our depth plus image model directly to this 4032px$\times$3024px$\times$3 channel$\times$42 frame volume.


\section{Unsupervised Depth Estimation}
In this section, we propose a method for depth estimation from long-burst data. We first lay out the projection model our method relies on, before introducing the scene model, loss functions, and training procedure used to optimize it.

\vspace{0.5em}
\noindent\textbf{Projection Model.}\hspace{0.1em} Given an image stack $I(u,v,\textsc{n})$, where $u,v\in[0,1]$ are continuous image coordinates and $\textsc{n}\in[0,1,...41]$ is the frame number, we aim to {condense the information in $I(u,v,\textsc{n})$ to a single compact projection model}. Given that the motion between frames is small, and image content is largely overlapping, we opt for an RGB-D representation which models each frame of $I(u,v,\textsc{n})$ as the deformation of some \emph{reference} image $I(u,v)$ projected through depth $D(u,v)$ with a change in camera pose $P(\textsc{n})$. We expand this process for a single point at coordinates $u,v$ in the reference frame. Let
\begin{equation}\label{eq:sample_img}
C = [R,G,B]^\top = I(u,v),\quad d = D(u, v)
\end{equation}
be a sampled colored point $C$ at depth $d$. Before we can project this point to new frame, we must first convert it from camera $(u,v)$ to world $(x,y,z)$ coordinates. We assume a pinhole camera model to \emph{un-project} this point via
 \vspace{-0.2em}
\begin{equation}\label{eq:unprojection}
\left[\arraycolsep=2.0pt
\begin{array}{c}
x \\
y \\
z \\
1
\end{array}\right]
= \bm{\pi}^{-1}\left(
\left[\arraycolsep=2.0pt
\begin{array}{c}
u \\
v \\
d \\
\end{array}\right]
;K\right)
=
\left[\arraycolsep=2.0pt
\begin{array}{c}
d(u-c_x) / f_x \\
d(v-c_y) / f_y \\
d \\
1
\end{array}\right],
\end{equation}
 \vspace{-0.2em}
where $K$ are the corresponding camera intrinsics with focal point $(f_x, f_y)$ and principal point $(c_x, c_y)$. We transform this point from the reference frame to target frame $\textsc{n}$, with camera pose $P(\textsc{n})$, via
 \vspace{-0.2em}
\begin{equation}\label{eq:pose}
\left[\arraycolsep=2.0pt
\begin{array}{c}
x^{\textsc{n}} \\
y^{\textsc{n}} \\
z^{\textsc{n}} \\
\end{array}\right]
=
\left[R(\textsc{n})\,|\,T(\textsc{n}) \right]
\left[\arraycolsep=2.0pt
\begin{array}{c}
x \\
y \\
z \\
1
\end{array}\right]
=
\left[P(\textsc{n})\right]
\left[\arraycolsep=2.0pt
\begin{array}{c}
x \\
y \\
z \\
1
\end{array}\right].
\end{equation}
\begin{figure}[t]
    \centering
    \includegraphics[width=\linewidth]{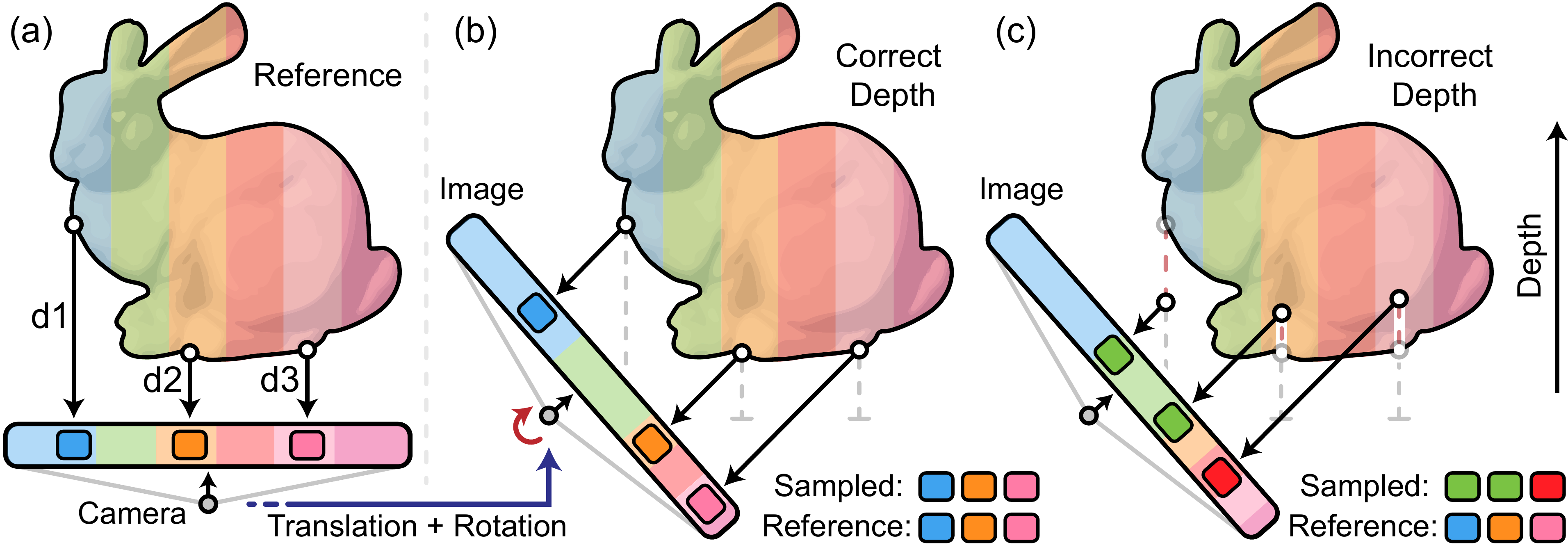}
    \caption{A 2D example of reprojection and sampling. When a reference view (a) is projected to new view with known camera rotation and translation, if the points' depths are accurately estimated they project to sample matching colors in the new image (b). If depths are inaccurate, as in (c), they do not sample corresponding colors, and instead incur \emph{photometric loss}.}
    \label{fig:raw_projection}
    \vspace{-1em}
\end{figure}
\noindent Here, $P(\textsc{n})$ is decomposed into a $3{\times}3$ rotation matrix $R(\textsc{n})$ and $3{\times}1$ translation vector $T(\textsc{n})=\left[t_x, t_y, t_z\right]^\top$. Reverse of the process in Eq.~\eqref{eq:unprojection}, we now \emph{project} this point from the world coordinates $(x^\textsc{n},y^\textsc{n},z^\textsc{n})$ in  frame $\textsc{n}$ to camera coordinates $(u^\textsc{n},v^\textsc{n})$ in the same frame as
 \vspace{-0.2em}
\begin{equation}\label{eq:projection}
\left[\arraycolsep=2.0pt
\begin{array}{c}
u^\textsc{n} \\
v^\textsc{n} \\
\end{array}\right]
= \bm{\pi}\left(
\left[\arraycolsep=2.0pt
\begin{array}{c}
x^{\textsc{n}} \\
y^{\textsc{n}} \\
z^{\textsc{n}} \\
\end{array}\right]
;K(\textsc{n})\right)
=
\left[\arraycolsep=2.0pt\def\arraystretch{1.2}
\begin{array}{c}
(f^\textsc{n}_{x} x^\textsc{n}) / z^\textsc{n} + c^\textsc{n}_{x}\\
(f^\textsc{n}_{y} y^\textsc{n}) / z^\textsc{n} + c^\textsc{n}_{y}
\end{array}\right],
\end{equation}
\begin{figure*}[t]
 \vspace{-0.9em}
    \centering
    \includegraphics[width=\linewidth]{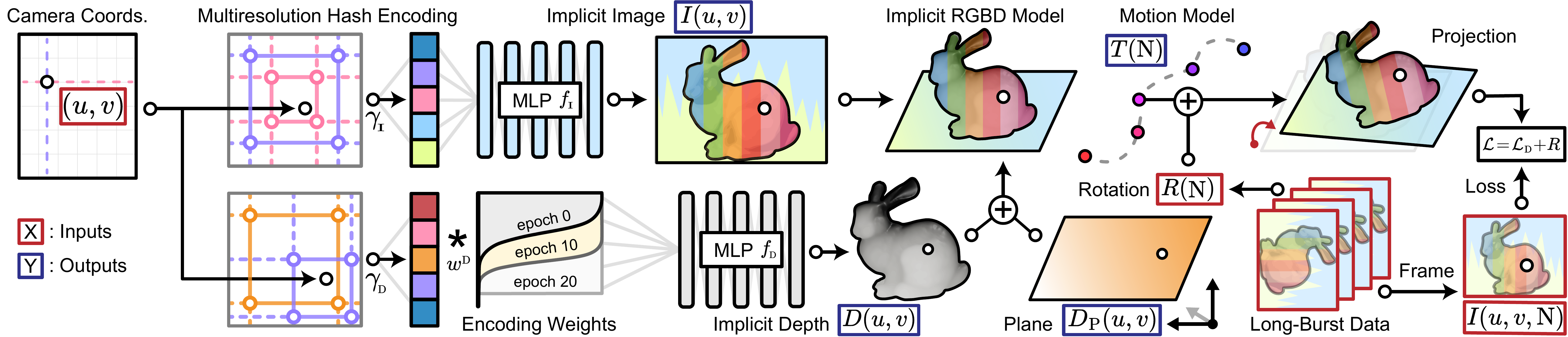}
    \caption{We model a long-burst capture as a single, fully-differentiable forward model comprised of an implicit image $I(u,v)$ projected through implicit depth $D(u,v)$ with motion model $[R(\textsc{n})|T(\textsc{n})]$. Calculating reprojection loss with respect to the the captured image stack $I(u,v,\textsc{n})$, we train this model end-to-end and distill high quality depth and camera motion estimates directly from the burst data.}
    \label{fig:method}
    \vspace{-1em}
\end{figure*}
\noindent where $K(\textsc{n})$ are the frame intrinsics. We can now use these coordinates to sample a point from the full image stack
\begin{equation}\label{eq:sample_volume}
C^\textsc{n} = I(u^\textsc{n},v^\textsc{n},\textsc{n}), \quad \mathcal{L}_{photo} = |C - C^\textsc{n}|.
\end{equation}
Here $\mathcal{L}_{photo}$ is \emph{photometric loss}, the difference in color between the point we started with in the reference frame and what we sampled from frame $\textsc{n}$. Given ideal multiview imaging conditions -- no occlusions, imaging noise, or changes in scene lighting -- if depth $d$ and pose change $P(\textsc{n})$ are correct, we will incur no photometric loss $\mathcal{L}_{photo}\,{=}\,0$ as we sample matching points in both frames. This is visualized in Fig.~\ref{fig:raw_projection}. \emph{Inverting this observation}, we can solve for unknown $D(u,v)$ and $P(\textsc{n})$ by finding ones that \emph{minimize photometric loss}~\cite{schonberger2016structure}.

\vspace{0.5em}
\noindent\textbf{Implicit Image Model.}\hspace{0.1em} In our problem setting, we are given a long-burst image stack $I(u,v,\textsc{n})$ and device rotation values $R(\textsc{n})$, supplied by an on-board gyroscope, and are tasked with recovering depth $D(u,v)$ and translation $T(\textsc{n})$ which make these observations consistent. Given the sheer number of pixels in $I(u,v,\textsc{n})$, in our case over \emph{500 million}, exhaustively matching and minimizing pixel-to-pixel loss is both computationally intractable and ill-posed. Under small camera motion, many depth solutions for a pixel can map it to identical-colored pixels in the image, especially in textureless parts of the scene. Traditional multi-view stereo (MVS) and bundle adjustment methods tackle this problem with feature extraction and matching~\cite{triggs1999bundle}, optimizing over a \emph{cost-volume} orders of magnitude smaller than the full image space. Here we strongly diverge from previous small motion works~\cite{im2015high,chugunov2022implicit,yu20143d}. Rather than divide the problem into feature extraction and matching, or extract features at all, we propose a single fully differentiable forward model trained \emph{end-to-end}. Depth is distilled as a product of fitting this \emph{neural scene model} to long-burst data. We start by redefining $I(u,v)$ from a static reference image to a learned implicit representation
\begin{align}\label{eq:implicit_img}
    I(u,v) &= f_\textsc{i}(\gamma_\textsc{i}(u,v;\; \mathrm{params_{\gamma\textsc{i}}});\;\theta_\textsc{i}) \nonumber\\ 
    \mathrm{params_{\gamma\textsc{i}}} &= \left\{N_{min}^{\gamma\textsc{i}},\, N_{max}^{\gamma\textsc{i}},\,L^{\gamma\textsc{i}},\, F^{\gamma\textsc{i}},\,T^{\gamma\textsc{i}}\right\} 
\end{align}
where $f_\textsc{i}$ is a multi-layer perceptron (MLP)~\cite{hornik1989multilayer} with learned weights $\theta$. This MLP learns a mapping from $\gamma_\textsc{i}(u,v)$, a positional encoding of sampled camera coordinates, to image color. Specifically, we borrow the multi-resolution hash encoding from Müller et al.\cite{muller2022instant} for its spatial aggregation properties. The parameters in $params_{\gamma\textsc{i}}$ determine the minimum $N_{min}^{\gamma\textsc{i}}$ and maximum $N_{max}^{\gamma\textsc{i}}$ grid resolutions, number of grid levels $L^{\gamma\textsc{i}}$, number of feature dimensions per level $F^{\gamma\textsc{i}}$, and overall hash table size $T^{\gamma\textsc{i}}$. 

\vspace{0.5em}
\noindent\textbf{Implicit Depth on a Plane Model.}\hspace{0.1em} Our depth model is a similar implicit representation with a \emph{learned planar offset}
\begin{align}\label{eq:depth}
    d &= D(u,v) = D_\textsc{p}(u,v) + f_\textsc{d}(\gamma_\textsc{d}(u,v; \mathrm{params}_{\gamma\textsc{d}}); \theta_\textsc{d})^+ \nonumber\\
    d_\textsc{p} &= D_\textsc{p}(u,v) = \mathrm{a}u + \mathrm{b}v + \mathrm{c},
\end{align}
where $\{\mathrm{a}, \mathrm{b}, \mathrm{c}\}$ are the learned plane coefficients, and $^+$ is the ReLU operation $max(0,x)$. Here $D_\textsc{p}(u,v)$ acts as the depth of the scene background -- the surface on or in front of which objects are placed -- which is often devoid of parallax cues. Then $f_\textsc{d}$ reconstructs the depth of the scene foreground content recovered from parallax in $I(u,v,\textsc{n})$. While it may seem that we are \emph{increasing} the complexity of the problem, as we now have to learn $I(u,v)$ in addition to $D(u,v)$, this model actually simplifies the learning task when compared to a static $I(u,v)$. Rather than solving for a perfect image from the get-go, $f_\textsc{i}$ can move between intermediate representations of the scene with blurry, noisy, and misaligned content, and is gradually refined during training.

\vspace{0.5em}
\noindent\textbf{Camera Motion Model.}\hspace{0.1em} Given the continuous, smooth, low-velocity motion observed in natural hand tremor~\cite{chugunov2022implicit}, we opt for a low-parameter B\'ezier curve motion model
\begin{align}\label{eq:spline_model}
    &T(\textsc{n})  = \mathrm{B}(\textsc{n}; \mathbf{P}^\textsc{t}, N^\textsc{t}_c), \; R(\textsc{n}) = R_{d}(\textsc{n}) + \eta_\textsc{r}\mathrm{B}(\textsc{n}; \mathbf{P}^\textsc{r}, N^\textsc{r}_c)\nonumber\\
    &\mathrm{B}(t;\mathbf{P}, N_c) = \sum_{i=0}^{N_c}
    \left(\arraycolsep=2.0pt
    \begin{array}{l}
    N_c \\
    \;i
    \end{array}
    \right)(1-t)^{N_c-i} t^i \mathbf{P}_i,
\end{align}
with $N_c$ number of control points $\mathbf{P}_i$. Translation estimates $T(\textsc{n})$ are learned from scratch, whereas rotations $R(\textsc{n})$ are initialized as device values $R_d(\textsc{n})$ with learned offsets weighted by $\eta_\textsc{r}$. Under the small angle approximation~\cite{im2015high}, we parameterize the rotational offsets $\mathbf{P^\textsc{r}}$ as 
\begin{equation}
    \mathbf{P^\textsc{r}_i} = \left[\begin{array}{ccc}
0 & -r^z & r^y \\
r^z & 0 & -r^x \\
-r^y & r^x & 0
\end{array}\right].
\end{equation}
The choice of $N_c$ controls the dimensionality of the curve on which motion lies -- e.g. $N_c\,{=}\,1$ restricts motion to be linear, $N_c\,{=}\,2$ is quadratic, and $N_c\,{=}\,42$ trivially overfits the data with a control point for each frame.

\vspace{0.5em}
\noindent\textbf{Loss and Regularization.}\hspace{0.1em} Putting all of the above together we arrive at the full forward model, illustrated in Fig.~\ref{fig:method}. Given that all of our operations -- from re-projection to B\'ezier interpolation -- are fully differentiable, \emph{we train all these components simultaneously, end-to-end, through stochastic gradient descent}. But to do this, we need an objective to minimize. We employ a weighted composite loss
\begin{align}
    \mathcal{L} &= \mathcal{L}_\textsc{d} + \alpha_\textsc{p}(\mathcal{L}_\textsc{p}/\mathcal{L}_\textsc{d})\mathcal{R}, \quad \alpha_\textsc{p}>0,\;\beta_\textsc{p}\geq1 \label{eq:loss1}\\
    \mathcal{L}_\textsc{d} &= |(C - C^\textsc{n}_\textsc{d})/(\mathrm{sg}(C) + \epsilon_\textsc{c})|^2 \label{eq:loss2}\\
    \mathcal{L}_\textsc{p} &= |(C - C^\textsc{p}_\textsc{d})/(\mathrm{sg}(C) + \epsilon_\textsc{c})|^2 \label{eq:loss3}\\
    \mathcal{R} &= |1 - d/d_\textsc{p}|^2 \label{eq:loss4}\\
    C &= I(u,v), \quad C^\textsc{n} = I(u,v,\textsc{n}), \quad C^\textsc{n}_\textsc{p} = I(u,v,\textsc{n})_\textsc{p} \nonumber
\end{align}
Here $d$ is the depth output by our combined depth model, and $d_\textsc{p}$ is the depth of only the planar component as in Eq.~\eqref{eq:depth}. $C$ is a colored point sampled from our implicit image model, $C^\textsc{n}$ is the point sampled from the image stack $I(u,v,\textsc{n})$ following Eqs.~\eqref{eq:sample_img}\,--\,\eqref{eq:sample_volume} for depth $d=d$, and $C^\textsc{n}_\textsc{p}$ is the point sampled following Eqs.~\eqref{eq:sample_img}\,--\,\eqref{eq:sample_volume} for the plane depth $d=d_\textsc{p}$. The regularization term \eqref{eq:loss4} penalizes the magnitude of $f_\textsc{d}$, pulling the depth output towards the plane model. Losses \eqref{eq:loss2} and \eqref{eq:loss3} are relative square photometric errors between sampled colored points, where $\mathrm{sg}$ is the stop-gradient operator preventing the denominator $C$'s gradient from being back-propagated. This normalization by the approximate luminance of sampled points is effective in aiding the unbiased reconstruction of HDR images~\cite{lehtinen2018noise2noise}, and we refer the reader to derivations in Mildenhall et. al~\cite{mildenhall2022nerf} on its relation to tone-mapping. In \eqref{eq:loss1}, we combine the photometric loss term $\mathcal{L}_\textsc{d}$, which seeks to maximize overall image reconstruction quality, with a weighted regularization $R$ which penalizes divergence from the planar model. When  $\mathcal{L}_\textsc{p}\approx\mathcal{L}_\textsc{d}$ -- i.e. the depth offset from $f_d$ is not improving reconstruction quality over a simple plane -- the model is strongly penalized. As $\mathcal{L}_\textsc{d}$ decreases relative to $\mathcal{L}_\textsc{p}$ -- the implicit depth model $f_d$ improves reconstruction quality -- this penalty falls off. In this way, regions that are blurred, textureless, or otherwise have no meaningful parallax information are pulled towards a spatially consistent plane solution rather than spurious depth predictions from $f_d$. As otherwise, in these regions, there is no photometric penalty for incorrect and noisy depth estimates. The parameter $\alpha_\textsc{p}$ controls the strength of this regularization.

\begin{figure}[t]
    \centering
    \includegraphics[width=\linewidth]{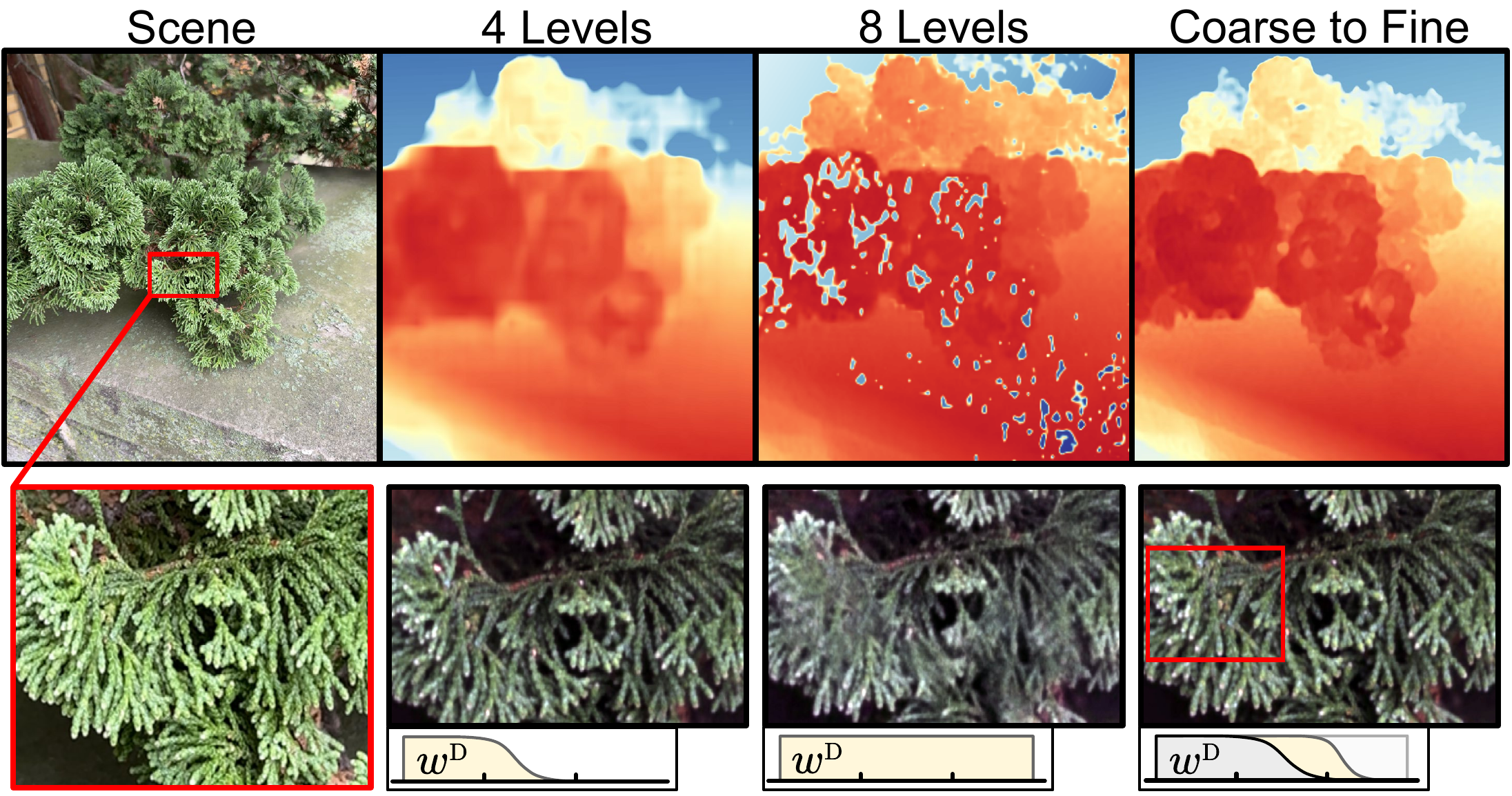}
    \caption{Ablation experiment on the effect of masked encoding levels. Using 4 encoding levels for depth leads to expected lower-resolution depth reconstructions. However, in \emph{8 Levels} where weights are not swept from coarse-to-fine, the reconstruction acquires sharp depth artifacts due to a positive feedback loop during training: high-frequency image gradients from $I(u,v)$ produce discontinuities in estimated depth $D(u,v)$, which in turn produce high-frequency images gradients in $I(u,v)$.}
    \label{fig:feature_masking}
    \vspace{-1.5em}
\end{figure}

\begin{figure*}[t]
 \vspace{-1.9em}
    \centering
    \includegraphics[width=\linewidth]{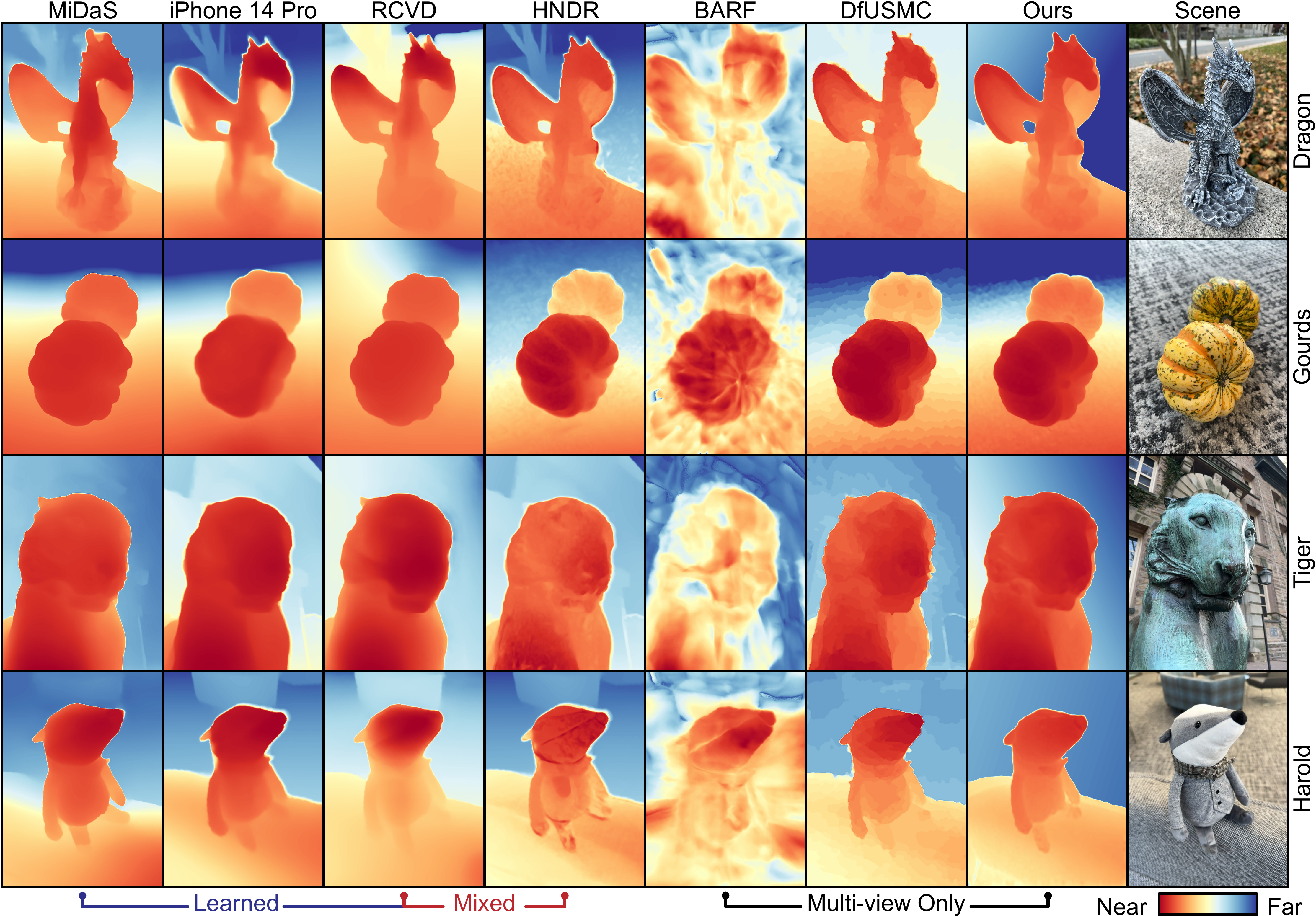}
    \caption{Qualitative comparison of reconstruction results of indoor and outdoor scenes for a range of learned, purely multi-view, and mixed depth estimation methods. Given the mix of depth scales, results are re-scaled by minimizing relative mean square error.}
    \label{fig:main_results}
    \vspace{-1em}
\end{figure*}

\vspace{0.5em}\noindent\textbf{Coarse-to-Fine Reconstruction.}\hspace{0.1em} First estimating low-resolution depths for whole objects before refining features such as edges and internal structures is a tried-and-true technique for improving depth reconstruction quality and consistency~\cite{chang2018pyramid,eigen2014depth}. However, one typical caveat of implicit scene representations is the difficulty of performing spatial aggregation -- an image pyramid is not well-defined for a continuous representation with no concept of pixel neighborhoods. Rather than try to aggregate outputs, we recognize that the multi-resolution hash encoding $\gamma_\textsc{d}(u,v)$ gives us control over the scale of reconstructed features. By masking the encoding $w^\textsc{d}\gamma_\textsc{d}(u,v)$ with weights $w^\textsc{d}_i\in[0,1]$ we can restrict the effective spatial resolution of the implicit depth network $f_d$, as two coordinates that map to the same masked encoding are treated as identical points by $f_d$. During training, we evolve this weight vector as
\begin{align}
    w^\textsc{d}_i &= 1/(1+exp(-ki))\nonumber\\
    k &= -k_{min} + (\mathrm{epoch} \cdot k_{max})/\mathrm{max\_epochs} 
\end{align}
which smoothly sweeps from passing only low-resolution grid encodings to all grid encodings during training, with $k_{min}$ and $k_{max}$ controlling the rate of this sweep. The effects of this masking are visualized in Fig.~\ref{fig:feature_masking}.

\vspace{0.5em}\noindent\textbf{Training and Implementation Details.}\hspace{0.1em} For simplicity of notation we have thusfar only worked with a single projected point. In practice, during a single forward pass of the model we perform \emph{one-to-all} projection of a batch of 1024 points at a time from the reference $I(u,v)$ to \emph{all 42 frames} in $I(u,v,\textsc{n}$). We perform stochastic gradient descent on $\mathcal{L}$ with the Adam optimizer~\cite{kingma2014adam}. Our implementation is built on tiny-cuda-dnn~\cite{muller2021real}, and on a single Nvidia A100 GPU has a training time of approximately $15$ minutes per scene. Our encoding parameters are $N_{min}^{\gamma\textsc{ i}}=8,N_{max}^{\gamma\textsc{i}}=2048,L^{\gamma\textsc{i}}=16,F^{\gamma\textsc{i}}=4,T^{\gamma\textsc{i}}=2^{22}$ and $N_{min}^{\gamma\textsc{ d}}=8,N_{max}^{\gamma\textsc{d}}=128,L^{\gamma\textsc{d}}=8, F^{\gamma\textsc{d}}=4,T^{\gamma\textsc{d}}=2^{14}$, as depth has significantly less high-frequency features than $I(u,v)$. The networks $f_\textsc{i}$ and $f_\textsc{d}$ are both 5-layer 128 neuron MLPs with ReLU activations. For the rotation offset weight we choose $\eta_\textsc{r}=10^{-4}$; regularization weight $\alpha_\textsc{p}=10^{-4}$ and $\epsilon_\textsc{c}=10^{-3}$; encoding weight parameters $k_{min}=-100, k_{max}=200$; and number of control points $N^\textsc{t}_c=N^\textsc{r}_c=21$, one for every two frames. We provide additional training details, and an extensive set of ablation experiments in the Supplemental Document to illustrate the effects of these parameters and how the above values were chosen. Our data capture app is built on the AVFoundation library in iOS 16 and tested with iPhone 12-, 13-, and 14-Pro devices. For consistency, a single 14 Pro device was used for all data captured in this work. RAW capture is hardware/API limited to ${\smallsim}21$~FPS, hence a two-second long-burst contains 42 frames.
\vspace{-0.5em}
\section{Assessment}\label{sec:results}
\begin{figure*}[t]
 \vspace{-1em}
    \centering
    \includegraphics[width=\linewidth]{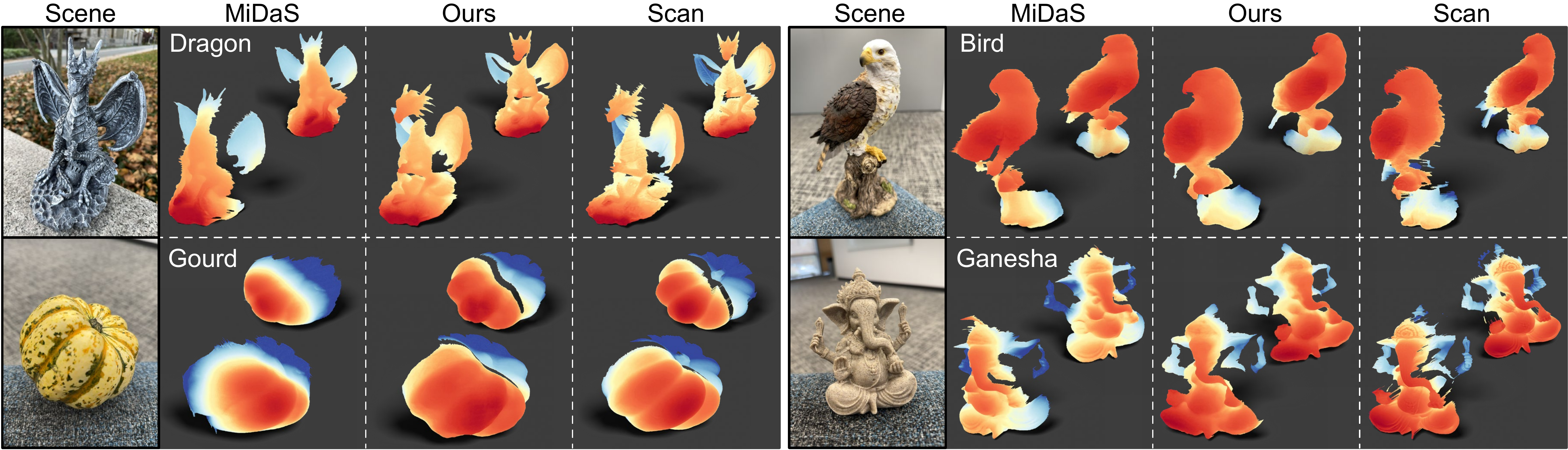}
    \resizebox{\linewidth}{!}{
     \begin{tabular}[b]{lccccccclccccccc}
     \midrule
          & MiDaS & iPhone & RCVD & HNDR & BARF & SfSM & Ours & & MiDaS & iPhone & RCVD & HNDR & BARF & DfUSMC & Ours\\
	\midrule
	\midrule
    \textit{Dragon} & .447$/$.224  & .482$/$.355 & .485$/$.237 & .470$/$.345 & .285$/$.339 & .166$/$.118 & \textbf{.129$/$.073} & \textit{Bird} & .283$/$.199  & .321$/$.282 & .322$/$.255 & .284$/$.274 & .172$/$.159 & .125$/$.161 & \textbf{.082$/$.058}    \\
    \textit{Gourd} & .233$/$.195  & .266$/$.229 & .235$/$.181 & .264$/$.225 & .821$/$.369 & .167$/$.141 & \textbf{.086$/$.078} & \textit{Ganesha} & .232$/$.194  & .283$/$.224 & .306$/$.239 & .275$/$.230 & .376$/$.328 & .132$/$.176 & \textbf{.094$/$.104}    \\
    \bottomrule
  \end{tabular}
  }
    \caption{Object reconstructions visualized as rendered meshes, with associated depth metrics formatted as \emph{relative absolute error / scale invariant error}. Edges over $10{\times}$ the length of their neighbors were culled to avoid connecting mesh features in occluded regions. }
    \label{fig:mesh_results}
    \vspace{-1em}
\end{figure*}

\noindent\textbf{Evaluation.} We compare our approach to the most similar purely multi-view methods BARF~\cite{lin2021barf} and Depth From Uncalibrated Small Motion Clip (DfUSMC)~\cite{ha2016high}, both of which estimate depth and camera motion directly from an input image stack. We note that BARF also has a similar implicit image generation model. We also compare to learned monocular methods: iPhone's 14 Pro's native depth output and MiDaS~\cite{ranftl2020towards}, a robust single-image approach. Lastly, we compare to Robust Consistent Video Depth Interpolation (RCVD)~\cite{kopf2021robust} and Handheld Multi-frame Neural Depth Refinement (HNDR)~\cite{chugunov2022implicit}, which both use multi-view information to refine initial depth estimates initialized from a learned monocular approach. The latter of which is most directly related to our approach as it targets close-range objects imaged with multi-view information from natural hand tremor, but relies on iPhone LiDAR hardware for depth initialization and pose estimation. All baselines were run on processed RGB data synchronously acquired by our data capture app, except for HNDR which required its own data capture software that we ran in tandem to ours. We note that other neural scene volume methods such as~\cite{mildenhall2022nerf} require COLMAP as a pre-processing step, which \emph{fails to find pose solutions for our long-burst data}. To assess absolute performance and geometric consistency, we scan a select set of complex 3D objects, illustrated in Fig.~\ref{fig:mesh_results}, with a commercial high-precision turntable structured light scanner (Einscan SP). We use this data to generate ground truth object meshes, which we register and render to depth with matching camera parameters to the real captures. For quantitative depth assessment, we use relative absolute error and scale invariant error, commonly used in monocular depth literature~\cite{Ummenhofer2017}; see the Supplemental Document for details.

\begin{figure}[t]
    \centering
    \includegraphics[width=\linewidth]{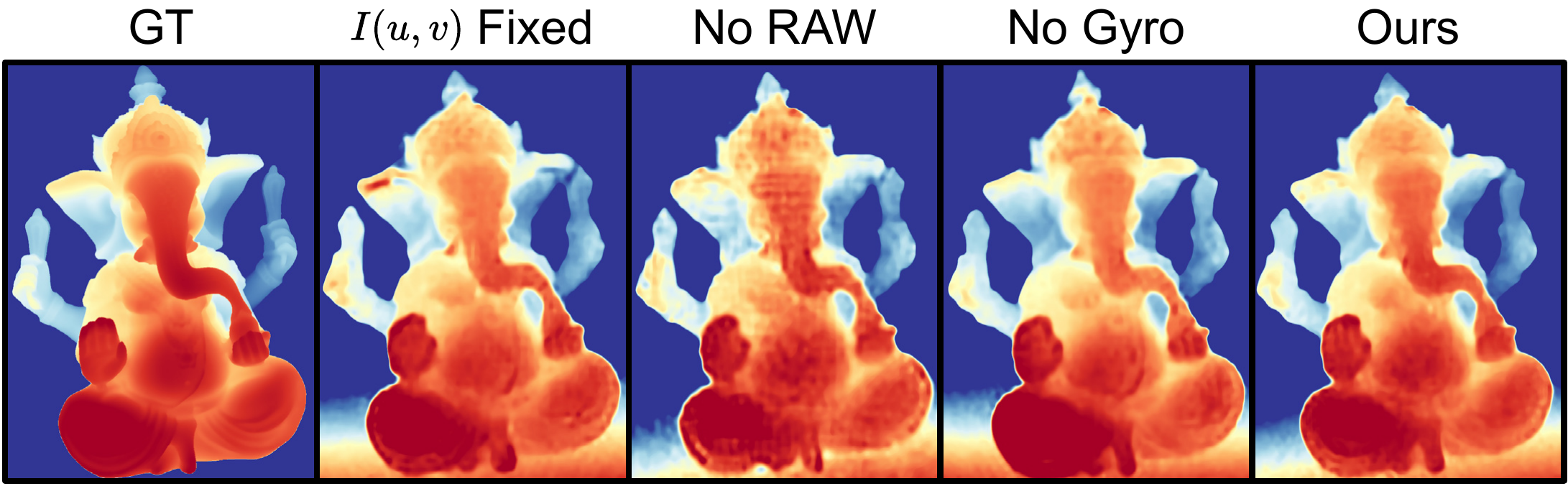}
        \resizebox{\linewidth}{!}{
     \begin{tabular}[b]{lcccclcccc}
     \midrule
          & Fixed & No RAW & No Gyro & Ours & &Fixed & No RAW & No Gyro & Ours \\
	\midrule
	\midrule
    Dragon & .150  & .142 & \textbf{.125} & .129 & Bird & .113 & .148 & .089 & \textbf{.082 } \\
     & $\overline{.081}$  & $\overline{.097}$ & $\overline{.080}$ & $\overline{\textbf{.073}}$ &  & $\overline{.099}$  & $\overline{.122}$ & $\overline{.078}$ & $\overline{\textbf{.058}}$  \\
     \hline
     Gourd & .091  & .093 & .088 & \textbf{.086} & Ganesha & .141  & .111 & .121 & .\textbf{094} \\
     & $\overline{\textbf{.078}}$  & $\overline{.081}$ & $\overline{.080}$ & $\overline{\textbf{.078}}$ &  & $\overline{.117}$  & $\overline{.104}$ & .$\overline{\textbf{094}}$ & $\overline{.104}$  \\
    \bottomrule
  \end{tabular}
  }
    \caption{Ablation study on the effects of fixing the image representation $I(u,v)$ to be the first long-burst frame $I(u,v,0)$, replacing the RAW data in $I(u,v,\textsc{n})$ with processed 8-bit RGB, or removing device initial rotation estimates from our model. Metrics formatted as \emph{relative absolute error / scale invariant error}.}
    \label{fig:rgb_comparisons}
    \vspace{-1.9em}
    
\end{figure}

\noindent\textbf{Reconstruction Quality.} Tested on a variety of scenes, illustrated in Fig.~\ref{fig:main_results}, we demonstrate high-quality object depth reconstruction outperforming existing learned, mixed, and multi-view only methods. Of particular note is how we are able to reconstruct small features such as \emph{Dragon}'s tail, \emph{Harold}'s scarf, and the ear of the \emph{Tiger} statue consistent to the underlying scene geometry. This is in contrast to methods such as RCVD or HNDR which either neglect to reconstruct the \emph{Tiger}'s ear or reconstruct it behind its head. Our coarse-to-fine approach also allows us to reconstruct scenes with larger low-texture regions, such as \emph{Harold}'s head, which produces striped depth artifacts for HNDR as it can only refine depth within a patch-size of high-contrast edges. Our depth on a plane decomposition avoids spurious depth solutions in low-parallax regions around objects, cleanly segmenting them from their background. This plane segmentation, and it's applications to image and depth matting, are further discussed in the Supplemental Document. In contrast to DfUSMC, which relies on sparse feature matches and RGB-guided filtering to in-paint contiguous depth regions, our unified end-to-end model \emph{directly} produces complete and continuous depth maps.

 In Fig.~\ref{fig:mesh_results}, we highlight our method's ability reconstruct complex objects. While from a \emph{single image} the learned monocular method MiDaS produces visually consistent depth results, from a \emph{single long-burst} our approach directly solves for geometrically accurate affine depth. This difference is most clearly seen in the \emph{Dragon} object, whose wings are reconstructed at completely incorrect depths by MiDAS, disjoint from the rest of the object. This improved object reconstruction is also reflected in the quantitative depth metrics, in which we outperform all comparison methods. Another note is that the most structurally similar method to ours, BARF -- which also learns an implicit image model and distills camera poses in a coarse-to-fine encoding approach -- fails to produce reasonable reconstructions. We suspect this is related to the findings of Gao et al.~\cite{gao2022monocular}, that NeRFs do not necessarily obey projective geometry during view synthesis for highly overlapping image content.

\noindent\textbf{Implicit Values of a Learned RAW Model.} In Fig.~\ref{fig:rgb_comparisons}, we observe the quantitative and qualitative effects of removing various key method components. For the \emph{No Gyro} tests we replace device rotations $R_d(\textsc{n})$ with identity rotation for all frames $\textsc{n}$ and learn offsets as usual. We find that while the use of a fixed reference image, 8-bit RGB, or no gyro measurements can reduce our model's average reconstruction quality, all these experiments still converge to acceptable depth solutions. This is especially true of the \emph{No Gyro} experiments, which for many scenes result in \emph{near identical} reconstructions. This further validates our model's ability to independently learn high quality camera pose estimates, and demonstrates its modularity with respect to input data and optimization settings -- applicable even to settings where RAW images and device motion data are not available.


\vspace{-0.5em}\section{Discussion and Future Work}
In this work, we demonstrate that from only a stack of images acquired during long-burst photography, with parallax information from natural hand-tremor, it is possible to recover high-quality, geometrically-accurate object depth.\\
\noindent\textbf{Forward Models.} Our static single-plane RGB-D representation could potentially be modified to include differentiable models of object motion, deformation, or occlusion. \\
\noindent\textbf{Image Refinement.} We use the learned image $I(u,v)$ as a vehicle for depth optimization, but it could be possible to flip this and use the learned depth $D(u,v)$ as a vehicle for aggregating RGB content (e.g., denoising or deblurring).
\noindent\textbf{From Pixels to Features.} Low-texture or distant image regions have insufficient parallax cues for ray-based depth estimation. Learned local feature embeddings could help aggregate spatial information for improved reconstruction.

\noindent\textbf{Acknowledgements.} We thank Jinglun Gao and Jun Hu for their support in developing the data capture app. Ilya Chugunov was supported by NSF GRFP (2039656). Felix Heide was supported by a Packard Foundation Fellowship, an NSF CAREER Award (2047359), a Sloan Research Fellowship, a Sony Young Faculty Award, a Project X Innovation Award, and an Amazon Science Research Award.

{\small
\bibliographystyle{ieee_fullname}
\bibliography{cvpr}
}

\vspace{1em}
\section{Supplemental Document}
In this document we provide supporting material including additional results, implementation details, ablation experiments, and further analysis in support of the findings from the main text. We organize this material as follows:
\begin{itemize}
\itemsep0em 
    \item Section~\ref{sec:implementation}: Data, training, and evaluation details.
    \item Section~\ref{sec:ablation}: Additional ablation studies on manipulating network and encoding parameters.
    \item Section~\ref{sec:additional_results}: Additional reconstruction results and examination of challenging imaging settings. 
    \item Section~\ref{sec:synthetic}: Additional validation on simulated data with evaluation of motion estimates.
    \item Section~\ref{sec:experiments}: Applications to depth and image matting. 
\end{itemize}
\setcounter{figure}{0}
\renewcommand*{\theHfigure}{chX.\the\value{figure}}
\setcounter{section}{0}
\renewcommand*{\theHsection}{chX.\the\value{section}}
\renewcommand\thesection{\Alph{section}}
\section{Implementation Details}
\label{sec:implementation}
\begin{figure}[t!]
    \centering
    \includegraphics[width=\linewidth]{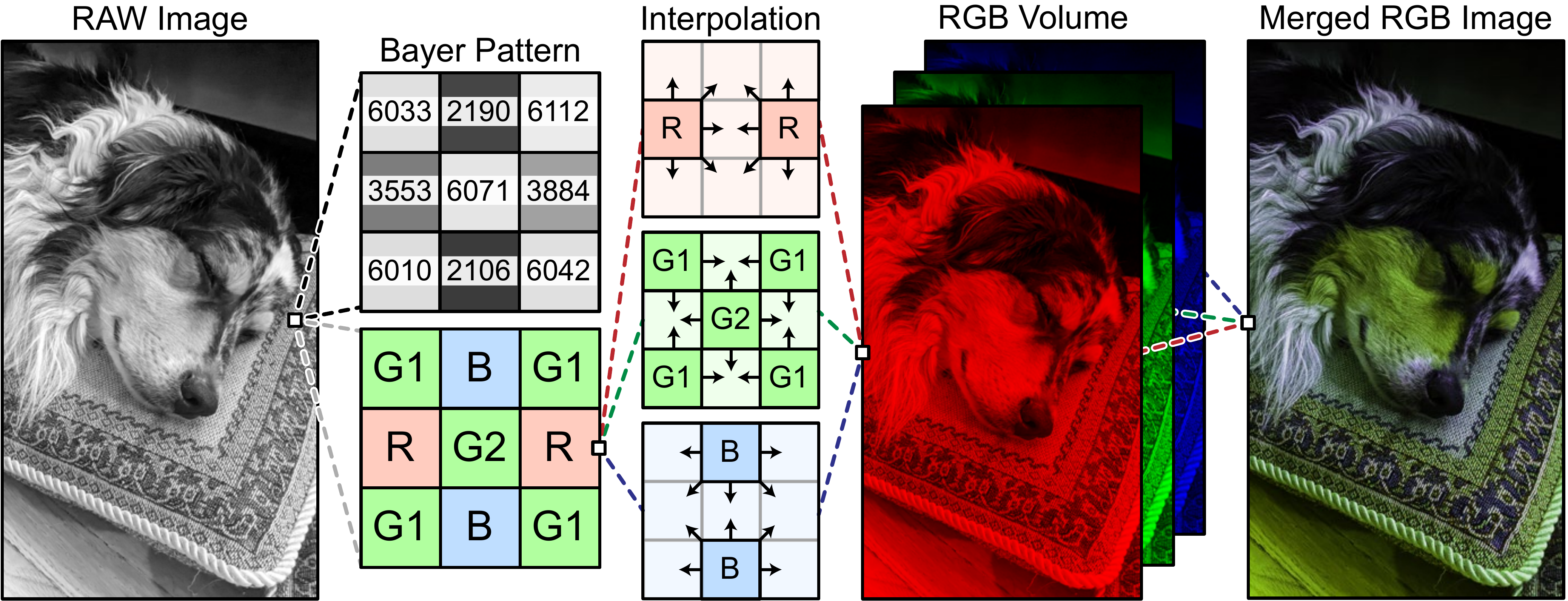}
    \caption{The bayer color filter array on a camera sensor produces a spatially "mosaicked" RAW image, where each 2$\times$2 block contains a blue, red, and two green pixels. Rather than mix channel content to "demosaick" the image, we separate these channels into three planes and only linearly interpolate gaps between measured pixels, preserving the original RAW values.}
    \label{fig:bayer_array}
    \vspace{-1em}
\end{figure}

\begin{figure}[t!]
    \centering
    \includegraphics[width=\linewidth]{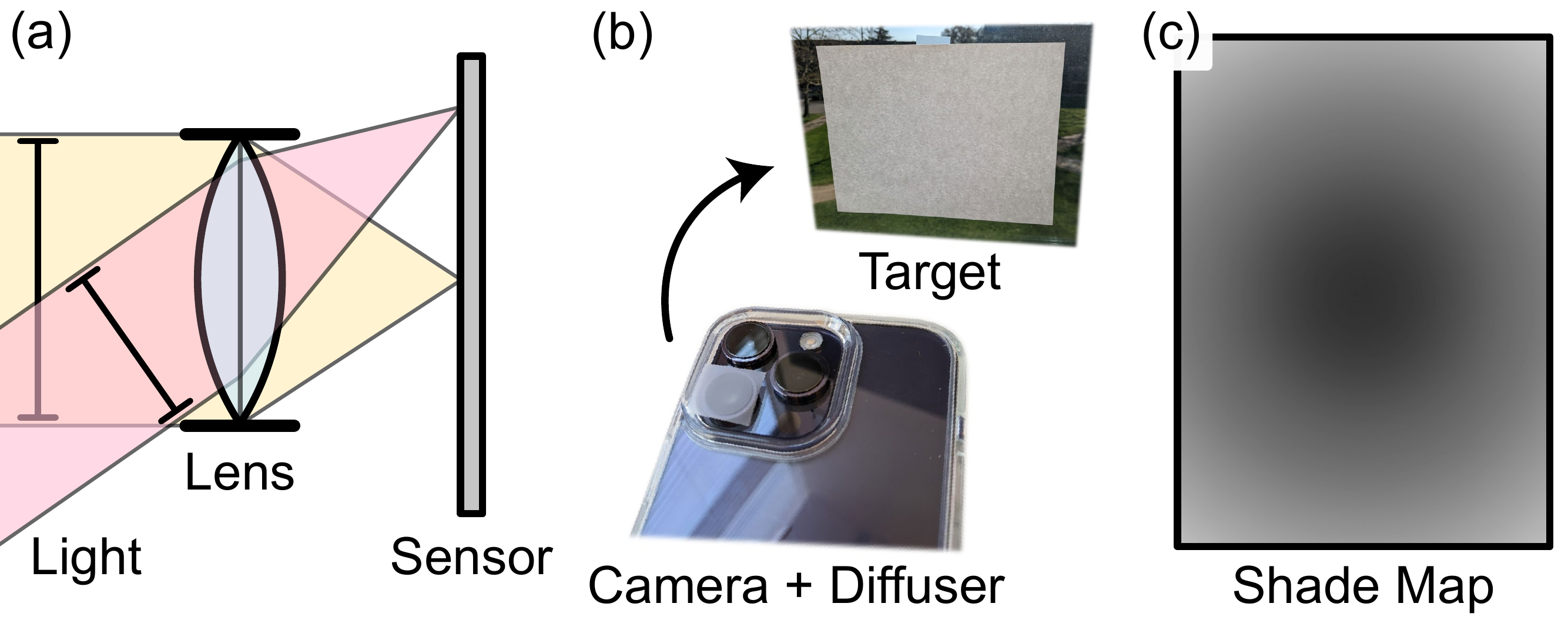}
    \caption{(a) Lens shading is an effect caused by the geometry of the camera lens assembly, where regions close to the edge of the image sensor receive less total light than the center. (b) Camera manufacturers calibrate for this by capturing what should be a uniformly bright scene and (c) generating a \emph{shade map} to compensate for the observed fall-off in brightness in the image.}
    \label{fig:lens_shading}
    \vspace{-1em}
\end{figure}

\noindent\textbf{Long-Burst Data.} We acquire long-burst data through a custom app built on the AVFoundation camera framework for iOS 16. While the vanilla AVFoundation framework offered a default method to capture burst or bracketed sequences, it was limited to only four frame sequences with significant overhead between captures, necessitating custom streaming code to save a longer continuous sequence of RAW data. A restriction we could not lift, however, is the inability to stream RAW captures from multiple cameras simultaneously. If this were possible, one could potentially use parallax and focus cues between two synchronized camera streams -- for example the wide and ultra-wide cameras -- to further improve reconstruction in the overlap of their fields of view. During capture, we record the following: Bayer CFA RAWs (42 frames $4032{\times}3024$px), processed RGB images (42 frames $1920{\times}1440$px), depth maps (42 frames $320{\times}240$px), frame timestamps, ISO, exposure time, brightness estimates, black level, white level, camera intrinsics, lens distortion tables, device acceleration estimates (${\smallsim}200$ measurements at 100Hz), device rotation estimates (${\smallsim}200$), and motion data timestamps (${\smallsim}200$). To preserve measured RAW values, we convert the single channel Bayer CFA images to three channel RGB volumes as shown in Fig.~\ref{fig:bayer_array}, linearly interpolating to fill missing values. To account for lens shading effects in bright scenes we also estimate a shade map with the help of a simple diffuser and uniform light source, illustrated in Fig.~\ref{fig:lens_shading}. We note that neglecting to compensate for lens shading can disrupt depth estimation in the corners of the image as matching pixels no longer have uniform brightness between frames.

\vspace{0.6em}
\noindent\textbf{Training.} We sample 1024 points $(u,v)$ per iteration of training, projecting these to $42{\times}1024$ points in the image stack $I(u,v,\textsc{n})$, corresponding to 1024 points per frame. We perform 256 iterations per epoch, for 100 epochs of training with the Adam optimizer~\cite{kingma2014adam} with betas (0.9, 0.99) and epsilon $10^{-15}$. We exponentially decay learning rate during training with a factor of 0.98 per epoch. Training on a single Nvidia A100 takes approximately 15 minutes.

\begin{figure}[t]
    \centering
    \includegraphics[width=\linewidth]{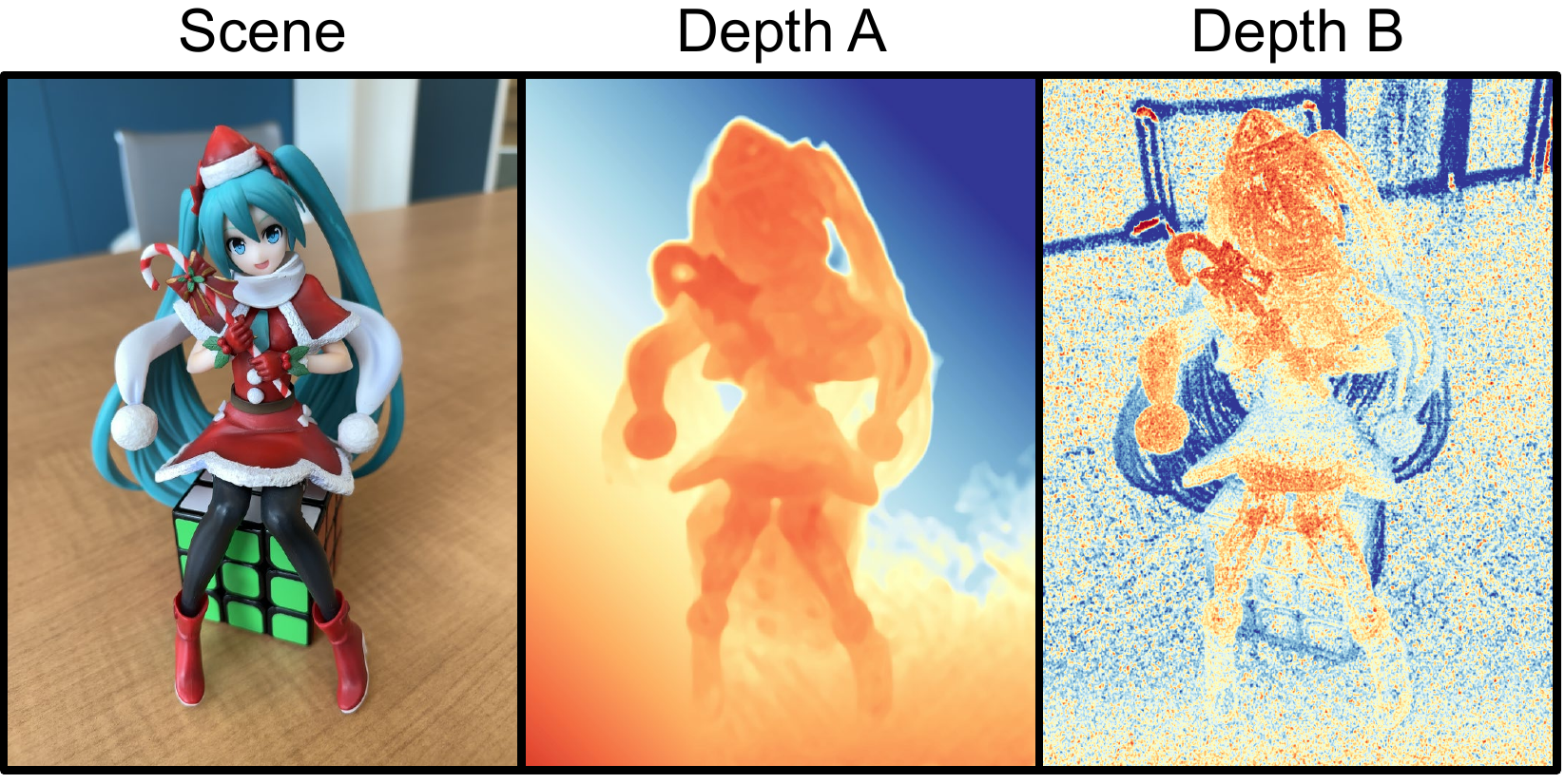}
    \caption{In this example both Depth A and B produce \emph{identical} reprojection error, but where Depth A models smooth geometry which warps the image between frames to model parallax, Depth B performs a brute-force mapping of individual pixels in the reference frame to points with similar values in the image stack.}
    \label{fig:reprojection_error}
    \vspace{-1em}
\end{figure}

\vspace{0.6em}
\noindent\textbf{Evaluation.} To generate depth maps we sample $D(u,v)$ at a grid of $(\textsc{h},\textsc{w})\,{=}\,(1920,1440)$ points $(u,v)\,{\in}\,[0,1]$. To reduce noise introduced by the stochastic training process we median filter this result with kernel size 13 before visualization. For depth evaluation, we use relative absolute error \emph{L1-rel} and scale invariant error \emph{sc-inv} metrics, that is
\begin{equation}
\text{L1-rel}(d, \hat d) = \frac{1}{\textsc{h}\textsc{w}}\sum_{u.v} \frac{|d(u,v) - \hat d(u,v)|}{\hat d(u,v)}\nonumber ,
\end{equation}
\noindent and
\begin{align}
\text{sc-inv}(d, \hat d) &= \sqrt{\frac{1}{\textsc{h}\textsc{w}} \sum_{u,v} \delta(u,v)^2-\frac{1}{(\textsc{h}\textsc{w})^2}(\sum_{u,v} \hat \delta(u,v))^2}\nonumber\\
\delta(u,v) &= \log(d(u,v)) - \log(\hat d(u,v)),\nonumber
\end{align}
which are often used in the monocular depth estimation literature~\cite{Ummenhofer2017} to compare approaches with varying scales and representations of depth. For methods such as MiDaS and RCVD we first convert inverse depth to depth before applying these metrics. We purposely avoid using photometric loss or reprojection error as comparison metrics~\cite{chugunov2022implicit} for similar arguments as discussed in Gao et al.~\cite{gao2022monocular}.
\begin{equation}
    \text{reprojection\_error}=\frac{1}{\textsc{h}\textsc{w}}\sum_{u,v,\textsc{n}} |I(u,v) - I(u^\textsc{n}, v^\textsc{n},\textsc{n})| \nonumber.
\end{equation}
Frames in a long-burst contain ${>}$90\% overlapping scene content, and so many non-physical solutions for depth will produce identical reprojection error as compared to more geometrically plausible depth maps. This is illustrated in Fig.~\ref{fig:reprojection_error}, where by ``tearing" the image -- compressing patches of similar colored pixels from the reference frame -- the non-physical depth incurs no additional photometric penalty, and so results in an identical reprojection error to a far more qualitatively plausible depth reconstruction.

\section{Additional Ablation Experiments}
\label{sec:ablation}
\begin{figure*}[t!]
    \centering
    \includegraphics[width=\linewidth]{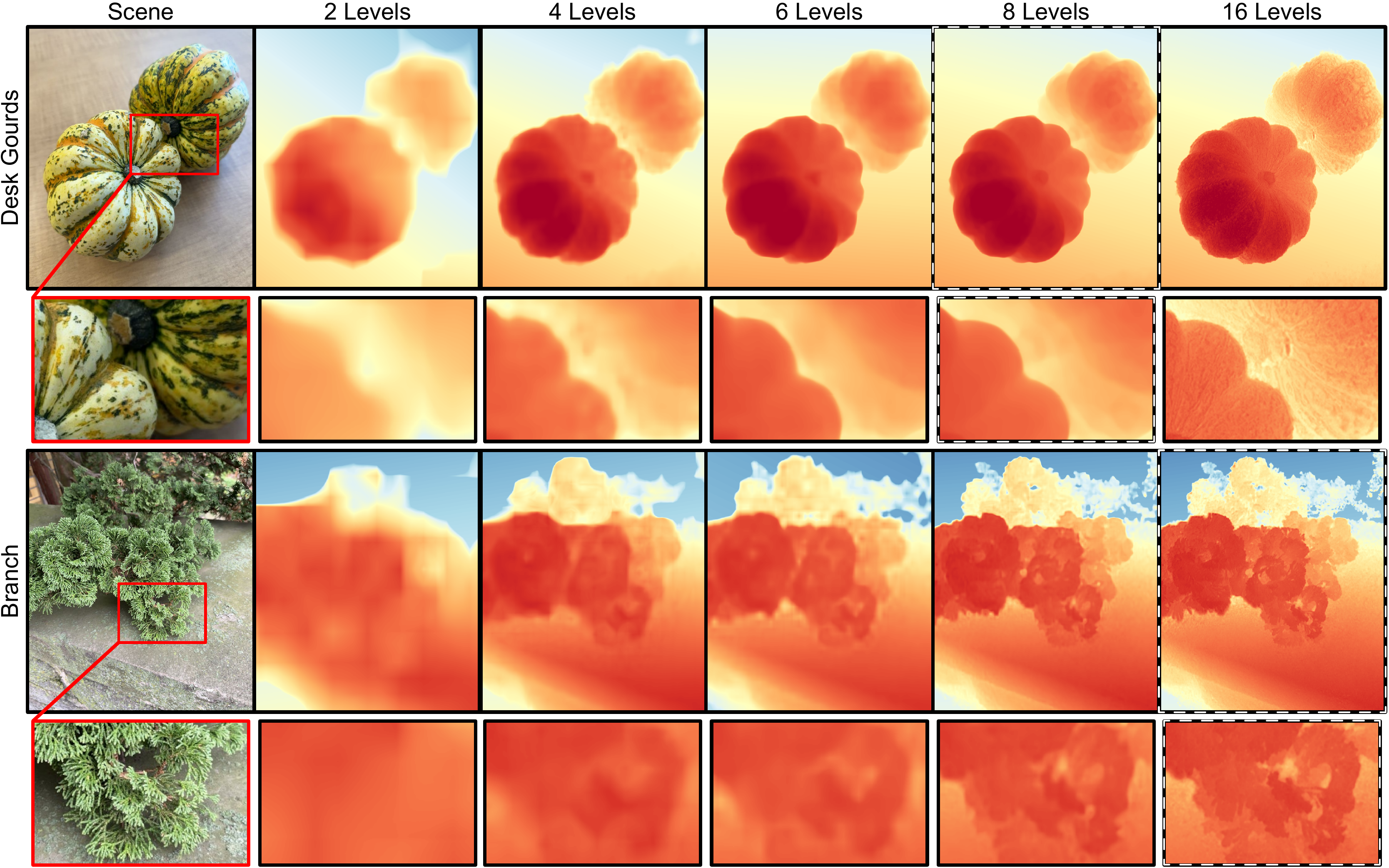}
    \caption{Ablation study on the effect of the number of levels $L^{\gamma\textsc{d}}$, and effective max resolution $N_{max}^{\gamma\textsc{d}}$, in the multiresolution hash encoding $\gamma_\textsc{d}$ on reconstruction. Here, given a scale factor of $\sqrt{2}$ between levels, $L^{\gamma\textsc{d}}=2,4,6,8,16$ correspond to $N_{max}^{\gamma\textsc{d}}=16,32,128,2048$. The qualitatively best reconstructions are highlighted with a dashed border.}
    \label{fig:encoding_layers}
\end{figure*}

\begin{figure*}[t!]
    \centering
    \includegraphics[width=\linewidth]{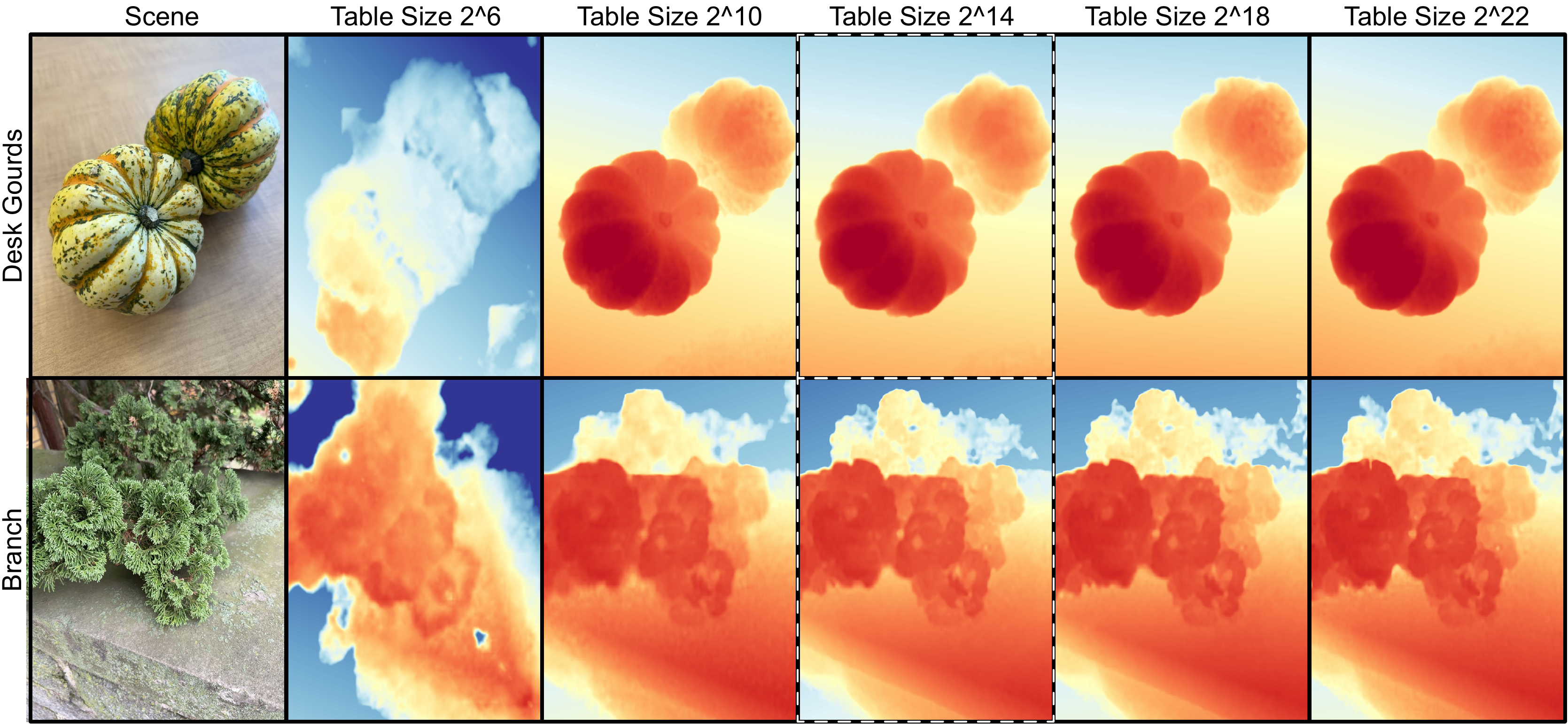}
    \caption{Ablation study on the effect of hash table size $T^{\gamma\textsc{d}}$ on reconstruction quality. Selected $T^{\gamma\textsc{d}}$ is highlighted with a dashed border.}
    \label{fig:encoding_hash}
\end{figure*}

\noindent\textbf{Encoding.} In this work we use the multiresolution hash encoding $\gamma_\textsc{d}$ to directly control what spatial information our implicit depth representation $f_\textsc{d}$ has access to during training. This in turn controls the scale of depth features we reconstruct, and presents a similar problem to choosing the scale factors in an image pyramid~\cite{adelson1984pyramid}. As we see in Fig.~\ref{fig:encoding_layers}, increasing the number of levels $L^{\gamma\textsc{d}}$ and effective max resolution $N_{max}^{\gamma\textsc{d}}$ increases the spatial frequency of reconstructed depth features. Scenes such as \emph{Branch} contain both high-frequency image and depth content, thin textured needles, and are best reconstructed by a fine resolution grid with $L^{\gamma\textsc{d}}=16$. The \emph{Desk Gourds}, however, have small image features in the patterns on the gourds, but relatively low-frequency depth features. Setting $L^{\gamma\textsc{d}}=16$ allows the network to overfit to these features and bleed image texture into the depth reconstruction. We select $L^{\gamma\textsc{d}}=8$ as a compromise between these imaging settings, but in practice, different scenes have different optimal encoding parameters for maximum reconstruction quality. We find hash table size $T^{\gamma\textsc{d}}$ significantly easier to tune, as choosing an overly large table size primarily affects model storage size, rather than reconstruction quality. We thus choose $T^{\gamma\textsc{d}}=2^{14}$, the smallest table size which does not lower the detail of depth reconstruction, as shown in Fig.~\ref{fig:encoding_hash}.

\begin{figure*}[htp!]
\vspace*{-1.2em}
    \centering
    \includegraphics[width=\linewidth]{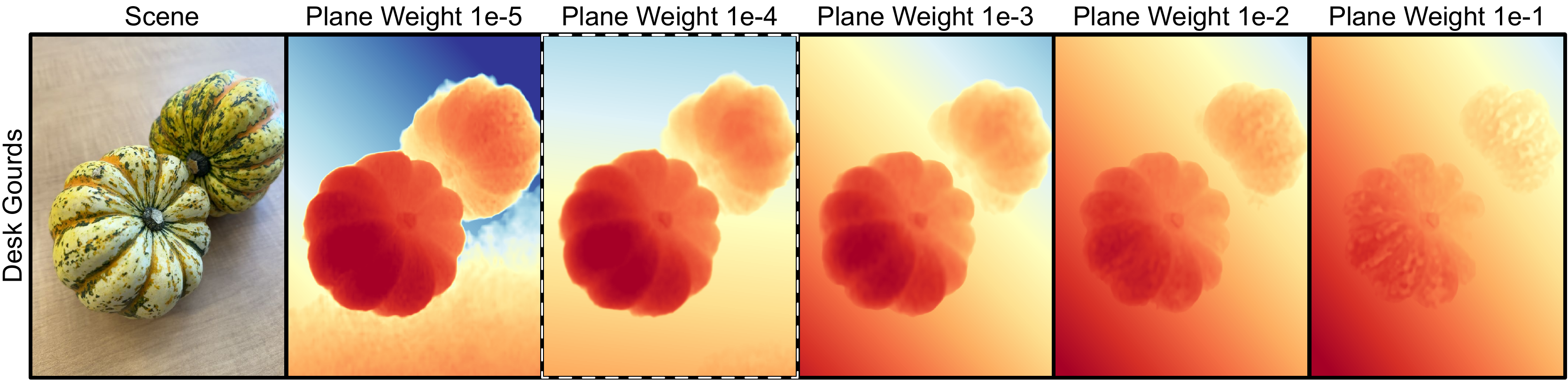}
    \caption{Ablation study on the effects of regularization weight $\alpha_\textsc{p}$ on reconstruction quality. Selected $\alpha_\textsc{p}$ highlighted with a dashed border.}
    \label{fig:plane_weight}
\bigskip
    \includegraphics[width=\linewidth]{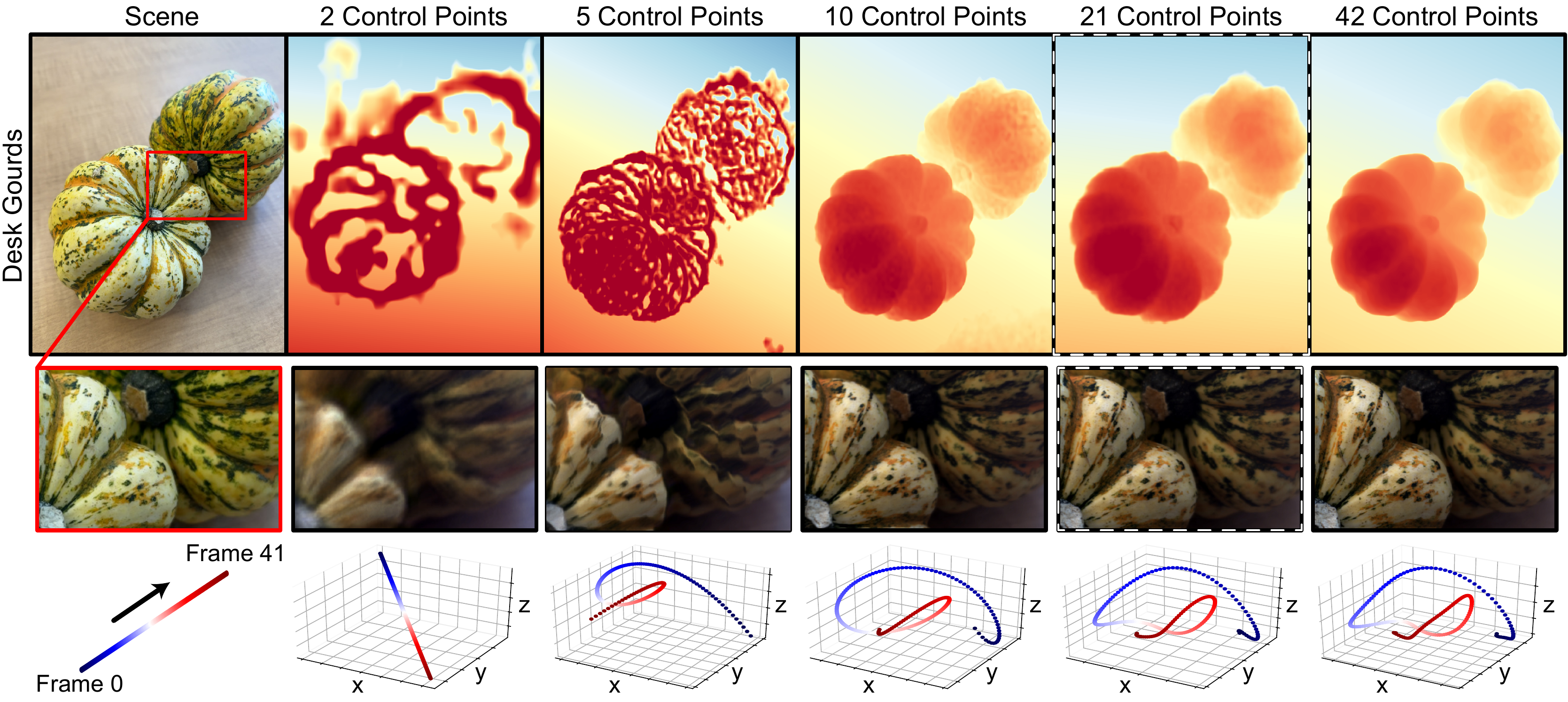}
    \caption{Ablation study on the effect of the number of chosen control points $N^\textsc{t}_c\,{=}\,N^\textsc{r}_c$ on reconstruction quality, with image reconstructions $I(u,v)$ and estimated motion paths plotted below. The selected number of control points is highlighted with a dashed border.}
    \label{fig:control_points}
\bigskip

    \includegraphics[width=\linewidth]{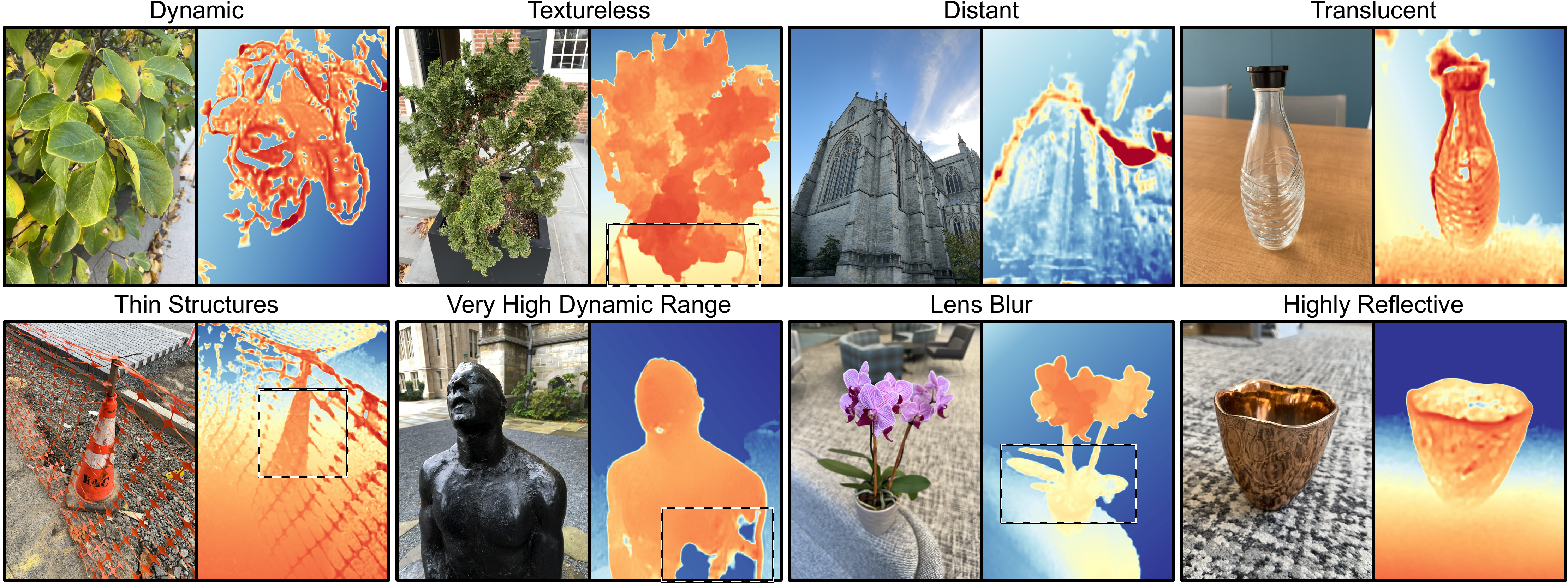}
    \caption{Depth reconstruction results for a set of challenging imaging scenarios. Not visible is the large motion of leaves in the \emph{Dynamic} scene, captured during high wind. Areas of interest are highlighted with a dashed border.}
    \vspace*{-0.5em}
    \label{fig:difficult_scenes}
\end{figure*}

\noindent\textbf{Depth Model.} The main adjustable parameter in our forward model is the plane regularization weight $\alpha_\textsc{p}$. This plane regularization affects depth reconstruction bidirectionally, regions with little parallax information are pulled towards the plane to remove spurious depth estimates, but in order to minimize depth offset, the plane is also pulled towards the reconstructed foreground objects. The effect of this can be seen in Fig.~\ref{fig:plane_weight}, where for very low $\alpha_\textsc{p}\,{\leq}\,10^{-5}$ this plane does not align with the foreground depth, and instead drifts into the background, causing a discontinuity in the reconstruction. Conversely, for large $\alpha_\textsc{p}\,{\geq}\,10^{-3}$, this regularization is so strong that the plane begins to cut into the foreground objects, flattening regions with low parallax information. We find $\alpha_\textsc{p}\,{=}\,10^{-4}$ to work well for a wide range of scenes, ``gluing'' the depth plane to the limit of reconstructed objects. We note that in scenes such as \emph{Desk Gourds}, and as we will see later with synthetic data, this plane accurately reconstructs the real geometry of the background. However, for many settings it is more akin to a \emph{segmentation mask} than depth, designating the area which we cannot reconstruct using parallax information.

\begin{figure*}[htp!]
    \centering
    \includegraphics[width=\linewidth]{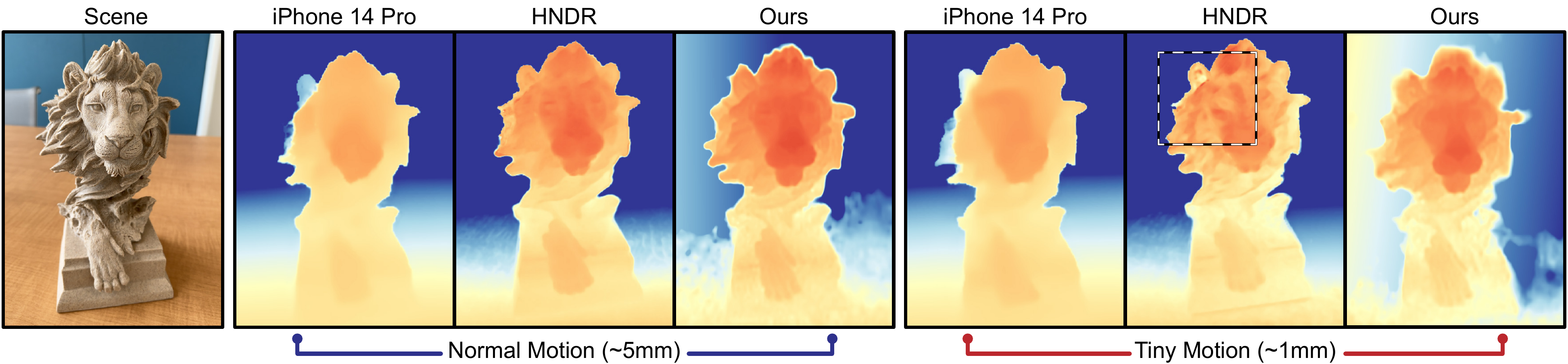}
    \caption{Depth reconstruction results for long-bursts captured with normal (approximately 5 millimeter maximum effective stereo baseline) and minimal (on the scale of a millimeter) hand shake motion. Major depth artifacts are highlighted with a dashed border.}
    \label{fig:small_motion}
\end{figure*}

\begin{figure}[t]
    \centering
    \includegraphics[width=\linewidth]{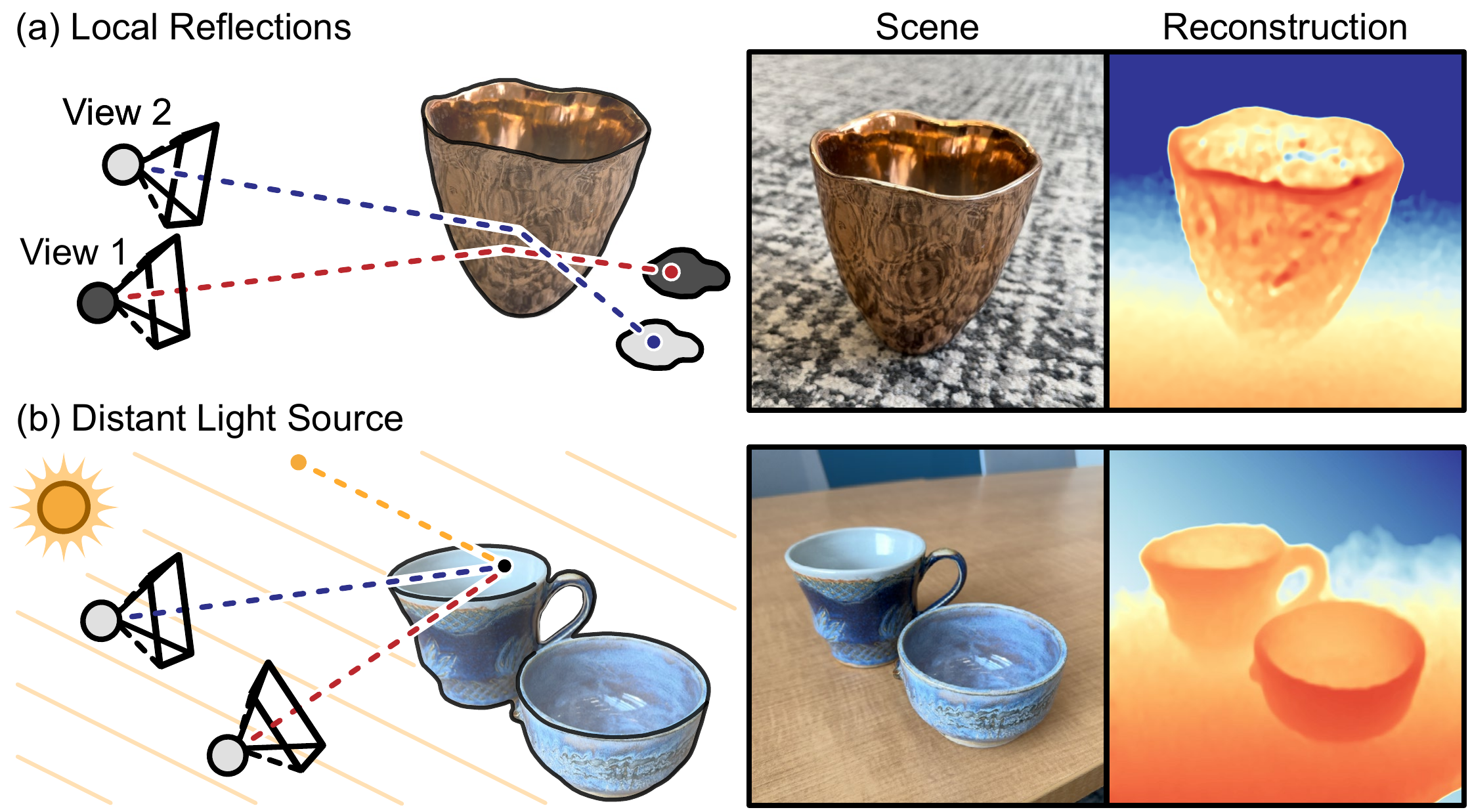}
    \caption{(a) Objects which reflect local scene content, in this example a mirror-finish copper pot reflecting the carpet around it, can completely break view consistency assumptions used for depth reconstruction. Even small view angle changes result in light paths which sample completely inconsistent colors in the surrounding environment. (b) In contrast, objects with specularities caused by a light source at effectively infinity, in this example polished ceramic reflecting sunlight, do not fully break view consistency.}
    \label{fig:non-lambertian}
    \vspace{-1em}
\end{figure}

\noindent\textbf{Motion Model.} We use a B\'ezier curve model to represent translation between frames, as natural hand-tremor draws a continuous low-velocity path during capture. By limiting the number of control points $N_c$ in this model we can enforce smoothness constraints on this motion, the effects of which are illustrated in Fig.~\ref{fig:control_points}. Not surprisingly, using too few control points does not allow us to faithfully model camera motion and results in blurry image reconstruction and inconsistent depth estimates. We thus choose the smallest number of control points which leads to successful image and depth reconstruction. We note that while for \emph{Desk Gourds} reconstruction succeeded with $N_c=42$, for many scenes setting $N_c\,{\geq}\,42$ leads to \emph{very unstable} training as the over-defined motion model can generate erratic high-velocity motion between frames.

\begin{figure*}[htp!]
    \centering
    \includegraphics[width=\linewidth]{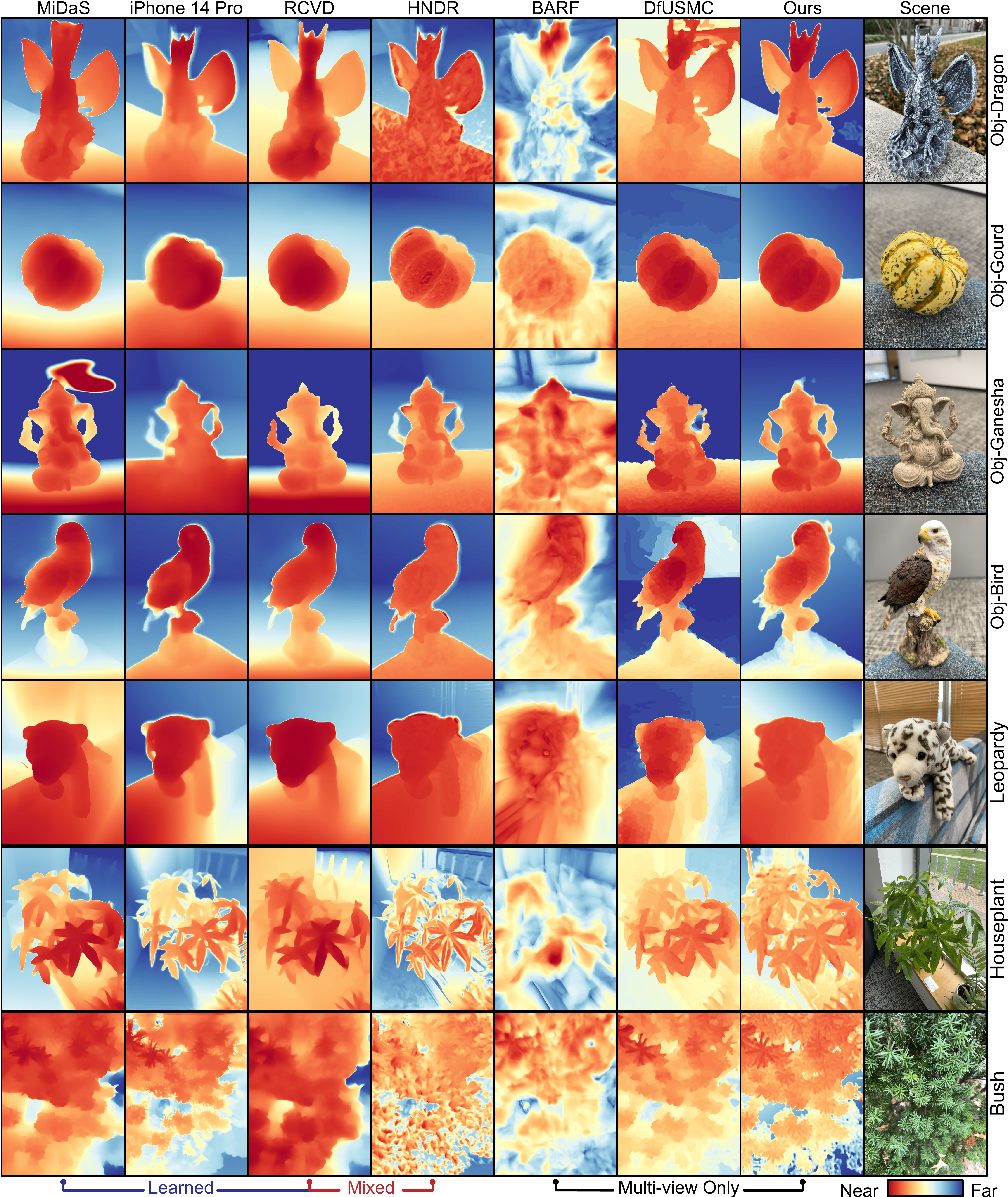}
    \caption{Reconstruction on 7 additional scenes for our method and a mix of learned, purely multi-view, and mixed depth estimation methods. Given the mix of depth representations, results are re-scaled by minimizing relative mean square error.}
    \label{fig:results_supp}
\end{figure*}

\section{Additional Reconstruction Results}
\noindent\textbf{Challenging Imaging Scenarios.} Given the fundamental building blocks of our approach, namely that it performs multi-view depth estimation through ray reprojection, some scenes will naturally be more difficult to reconstruct than others. As shown in Fig.~\ref{fig:difficult_scenes}, each of these scenarios presents its own set of challenges and direction of study. In the \emph{Dynamic} scene, we fail to reconstruct accurate depth for the majority of the plant leaves as they undergo deformation far larger than the parallax effects we observe in the long-burst. Our forward model has no way to model this deformation, and it is notoriously difficult to separate the effects of object motion from camera motion. The \emph{Textureless} and \emph{Distant} scenes present two sides of a similar problem, insufficient parallax information. While we are able to reconstruct the plant in \emph{Textureless}, the textureless planter provides no multi-view information from which to estimate depth except for along its edges, which we can track relative to the motion of the background. The church in \emph{Distant} is so far from the camera that it exhibits only fractions of pixel in disparity over the entire long-burst. In both these scenarios we need a mechanism to aggregate information in image space to make up for the lack of parallax. In \emph{Textureless} this would be in-painting the planters depth based on its edges, and in \emph{Distant} we would need to look at the deformation of larger image patches to estimate sub-pixel motion. The \emph{Thin Structures} reconstruction is partially successful, as in the foreground region we are able to track and reconstruct the depth of the thin orange mesh, but breaks down when it begins to overlap with the traffic cone. We suspect this is because our forward model is a single-layer RGB-D representation, with no explicit way to model for occlusions. In the region of the traffic cone is has to decide between reconstructing the cone or the mesh in front of it in the long-burst $I(u,v,\textsc{n})$ data, not both. Here, a layered depth representation could potentially solve this, but greatly increases the complexity of the problem as we would now need to learn an alpha map for each frame $\textsc{n}$ to sample these layers. For the \emph{Very High Dynamic Range} scene, we have specular reflections three orders of magnitude brighter than the shadowed portions of the statue. While using the fixed auto exposure and ISO settings we are able to reconstruct a large portion of the statue body with our RAW data. To reconstruct all the regions of the scene, including the dimly-lit body, our model could potentially be augmented to incorporate bracketed image data with varying exposure, similar to Mildenhall et al.~\cite{mildenhall2022nerf}, and perform joint HDR image volume and depth reconstruction. The \emph{Lens Blur} scene shows a loss in reconstruction quality due to portions of the scene being blurred by a shallow depth of field from the camera. Depth-from-defocus cues~\cite{xiong1993depth} could potentially help regularize reconstruction in these areas which are otherwise devoid of fine image features. Lastly, the \emph{Translucent} and \emph{Highly Reflective} settings both violate view consistency. Namely, changes in pixel colors can no longer be attributed solely to parallax or camera motion, and can be caused by seeing through or around the objects. We further discuss the reconstruction of non-lambertian objects in the next section.

\noindent\textbf{Non-Lambertian Reconstruction.}  While we focus on the reconstruction of primarily lambertian scenes -- matte, diffusely-reflective objects -- non-lambertian scenes provide an interesting set of both imaging challenges and opportunities. We first divide this setting into two categories: \emph{local reflections} and \emph{distant light sources}, illustrated in Fig.~\ref{fig:non-lambertian}. In the first setting, sampled light paths and colors can drastically change for even small view variations. As photometric matching tries to match reflected content, which does not follow the parallax motion of the reflective object itself, this produces erroneous depth estimates for objects such as the copper pot in Fig.~\ref{fig:non-lambertian} (a). With a distant light source, however, small changes in view angle result in the same apparent specularities as the path from the camera center to the illuminator remains connected. These specularities thus act as image texture, and exhibit the same parallax effects as the surface of the object. As seen in Fig.~\ref{fig:non-lambertian} (b) and the \emph{Tiger} scene, this does not disrupt depth reconstruction. This finding, that specularities from distant light sources act as object texture and local reflections do not, points towards an avenue of work in lighting separation and reflection removal. Regions which do not fit a static RGB-D model and incur large photometric penalties regardless of their depth, could be separated into view-dependent texture plus reflection components for later manipulation.

\noindent\textbf{Small Camera Motion.}  As hand shake is a naturally random process, long-burst captures have varying effective stereo baseline. While on average we can expect 5-6 millimeters of baseline~\cite{chugunov2022implicit}, if we are unlucky -- e.g. the user is not taking a breath and is rigidly holding the phone with two hands close to their body -- this motion can be as small as a millimeter. Illustrated in Fig.~\ref{fig:small_motion}, we see how our end--to-end camera pose estimation still converges in the minimal baseline setting, and how we are able to produce useable -- albeit blurrier -- depth estimates. This is in contrast to HNDR~\cite{chugunov2022implicit}, which uses the imperfect ARKit pose estimates for reprojection and produces major artifacts because of it, mapping incorrect pixel matches to spurious depths solutions. This demonstrates the value of continuous pose refinement, as mobile SLAM algorithms and COLMAP~\cite{schonberger2016structure} \emph{do not produce ground truth poses}.

\noindent\textbf{Additional Results.} \label{sec:additional_results}
Fig.~\ref{fig:results_supp} provides additional qualitative comparisons of our proposed approach to a wide set of baseline methods. This includes the four target objects used to demonstrate object reconstruction in the main text, prefixed with \emph{Obj-}. The visualizations also reflect the challenges in evaluating methods purely from depth maps, as geometric inconsistencies that are apparent in the mesh projections -- such as the distorted arms of \emph{Obj-Ganesha} -- are much harder to identify in these 2D visualizations. In addition to these objects, we include 3 scenes \emph{Leopardy, Bush, and Houseplant}, which demonstrate successful reconstruction with deceptive image features, small depth features, and large field of view respectively. Of particular note is how we are able to reconstruct the needles of the \emph{Bush} scene and individual leaves of \emph{Houseplant}, where other methods blend features at different depth levels together. 

\section{Synthetic Evaluation}
\label{sec:synthetic}
\begin{figure*}[ht!]
    \centering
    \includegraphics[width=\linewidth]{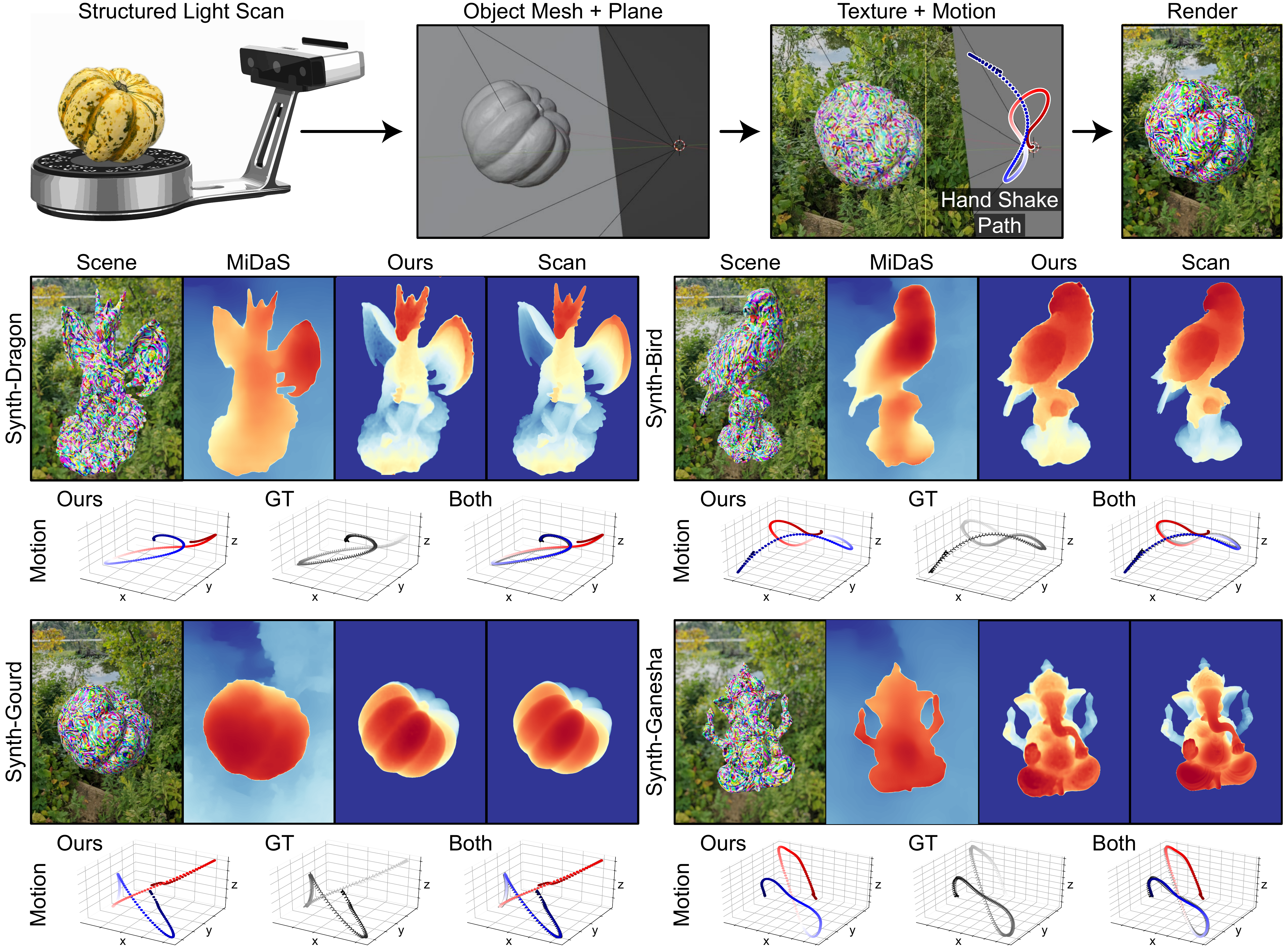}
    \caption{Depth reconstruction and motion estimation results for a set of simulated textured objects with realistic hand-tremor motion. Motion estimates are re-normalized and overlaid to demonstrate the accuracy in estimated camera trajectory to ground truth data.}
    \label{fig:synthetic_results}
\end{figure*}

\vspace{0.6em} \noindent\textbf{Setup.} To further validate our approach we use the high-fidelity structured light object scans we acquired for quantitative evaluation to generate simulated long-burst captures. Illustrated in Fig~\ref{fig:synthetic_results}, we apply a Voronoi color texture to the surface of these meshes, and place them in front of a tilted background plane with an outdoor image texture. We add depth-of-field effects and match camera intrinsics to our real captures -- using the ARKit poses captured by the software from Chugunov et al.~\cite{chugunov2022implicit} to generate realistic hand tremor motion paths -- and render frames at 16-bit color depth with Blender's Eevee engine. This synthetic data allows us to not only validate the fidelity of our object reconstructions, but also our estimated camera motion paths, for which we cannot otherwise get ground truth during ordinary captures.
\\

\noindent\textbf{Assessment on Synthetic Data.} We find that for this synthetic data -- in the absence of noise, lighting changes, and other imaging non-idealities -- we are able to recover nearly ground truth reconstructions of both the objects and background planes. This supports our plane plus offset depth model, which fits the simple plane to the out-of-focus background content instead of generating spurious depth estimates for regions without reliable parallax information. Though the colorful object textures make single-view depth estimation visually difficult, as illustrated by artifacts in the MiDaS reconstructions, these high-contrast cues allow our method to reconstruct even tiny features such as the tusks of the \emph{Synth-Ganesha}. This validates that even with small camera motion, given sufficient image texture we converge to geometrically correct solutions. In Fig.~\ref{fig:synthetic_results} we also see how the camera motion estimates converge close to ground truth as our method jointly refines depth and camera trajectory estimates during training. 

\section{Depth and Image Matting}
\label{sec:experiments}
\begin{figure*}[t!]
    \centering
    \includegraphics[width=\linewidth]{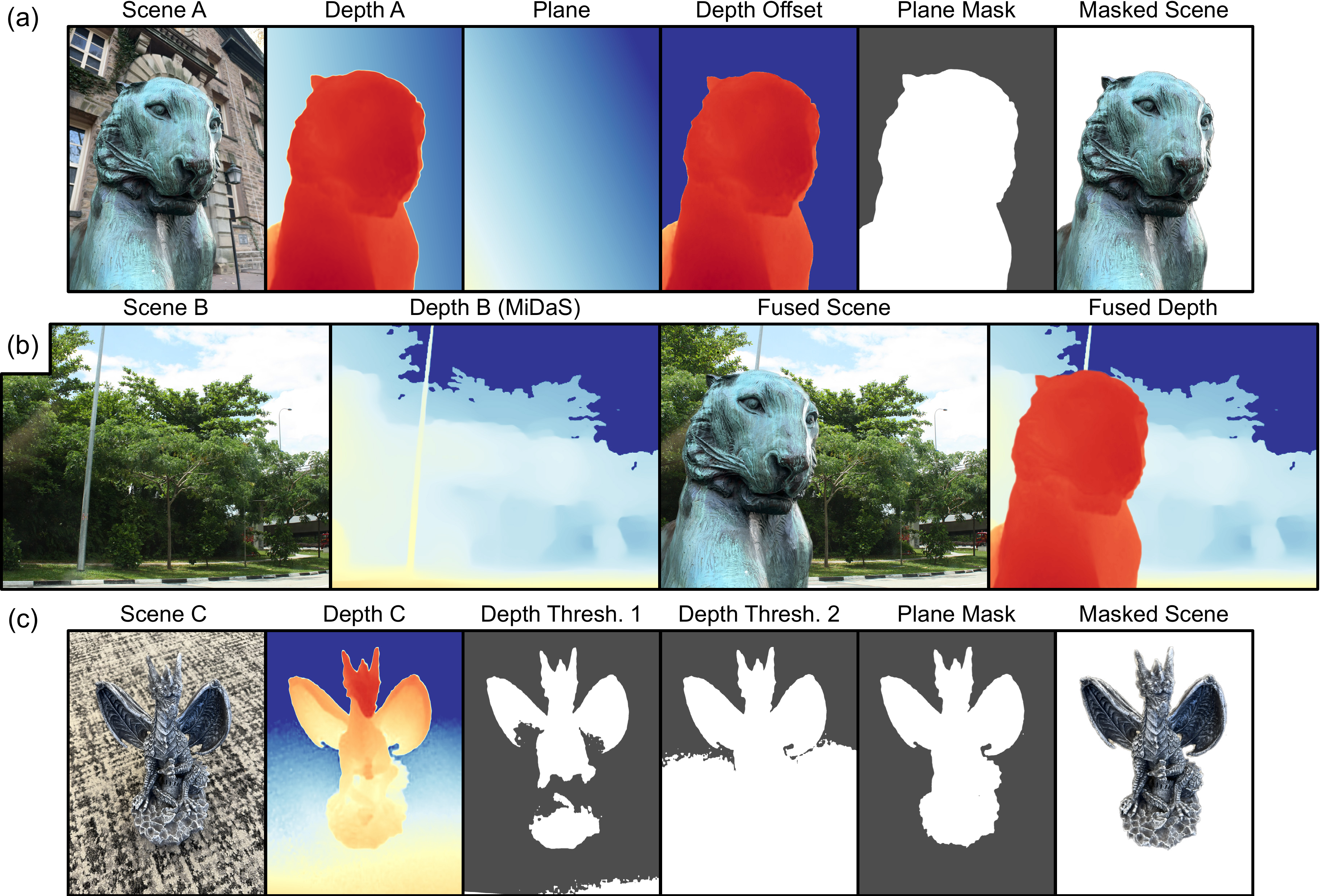}
    \caption{Image and Depth Matting. Example of scene editing enabled by our plane plus offset forward model. We can (a) threshold the depth offset component $d(u,v) - d_\textsc{p}$ to recover a mask of the object in focus and then (b) superimpose it over a new scene. (c) This works even for visually ambiguous scenes where simple depth thresholding fails.}
    \vspace*{5in}
    \label{fig:editting}
\end{figure*}

\noindent\textbf{Forward Model Decomposition.} In the proposed plane plus offset depth model, regions that do not generate sufficient parallax information are pulled towards the plane by the regularization term $R$. While we cannot recover meaningful depth from multiview in these regions, they prove \emph{useful for scene segmentation and editing}. Illustrated in Fig.~\ref{fig:editting} (a), by masking what parts of the image produce negligible depth offset, we are able to cleanly segment the tiger statue in \emph{Scene A} from its background. In Fig.~\ref{fig:editting} (b) we then superimpose this masked image over \emph{Scene B}, a separately captured tree-covered street. We run \emph{Scene B} through MiDaS to hallucinate the depth of the background trees, and overlay this with our geometrically-estimated depth of the tiger to produce a fused depth representation. In this way we leverage multiview information where we have it, and learned image priors where we do not. In Fig.~\ref{fig:editting} (c) we see an advantage of using this plane separation technique for segmentation over depth thresholding. As the floor under the dragon figure extends both in front of and behind the figure itself, setting a depth cutoff will always either miss a part of the figure, or include the area around it. Whereas as our plane here represents the depth of the floor, we can threshold the depth offset just like in Fig.~\ref{fig:editting} (a) to recover a \emph{high-quality mask of the object}. Thanks to being based on depth rather than image features, this approach has no problems with the visual ambiguity of the dragon and its background, which both contain high-frequency black and white textures.

\newpage



\end{document}